\documentclass[journal]{IEEEtran}
\usepackage{cite}
\usepackage{booktabs}
\usepackage{amsmath}
\usepackage{extarrows}
\usepackage[ruled]{algorithm2e}
\usepackage{makecell}
\usepackage{multirow}
\usepackage{color}
\usepackage{threeparttable}
\usepackage{soul, color, xcolor}
\soulregister\cite7 
\soulregister\eqref7
\soulregister{\ref}7 
\makeatletter
\renewcommand{\maketag@@@}[1]{\hbox{\m@th\normalsize\normalfont#1}}
\makeatother

\ifCLASSINFOpdf
  \usepackage[pdftex]{graphicx}
  \graphicspath{{../pdf/}{../jpeg/}}
  \DeclareGraphicsExtensions{.pdf,.jpeg,.png}
\else
  \usepackage[dvips]{graphicx}
  \graphicspath{{../eps/}}
  \DeclareGraphicsExtensions{.eps}
\fi

\begin{document}
%
\title{Hitchhiker: A Quadrotor Aggressively Perching on a Moving Inclined Surface Using Compliant Suction Cup Gripper}
%
%
%

\author{Sensen~Liu, 
        Zhaoying~Wang,
        Xinjun~Sheng$^{\ast}$,~\IEEEmembership{Member,~IEEE}
        and Wei~Dong,~\IEEEmembership{Member,~IEEE,}
\thanks{This work is partially supported by the Scientific and technical innovation 2030 -- ``Artificial Intelligence of New Generation" Major Project 
(2018AAA0102704) and National Natural Science Foundation of China (Grant No. 51975348). \textit{($^{\ast}$Corresponding author: Xinjun Sheng.)}}
\thanks{The authors are with the State Key Laboratory
of Mechanical System and Vibration, Robotics Institute, School of Mechanical and Engineering, Shanghai Jiao Tong University, 800
Dongchuan Road, Shanghai, China. \tt\footnotesize (e-mail: sensenliu@sjtu.edu.cn; dr.dongwei@sjtu.edu.cn; wangzhaoying@sjtu.edu.cn; xjsheng@sjtu.edu.cn).}}
%



\maketitle
\sethlcolor{yellow}
\begin{abstract}
Perching on {the surface} of moving objects, like vehicles, could extend the flight {time} and  range of quadrotors. Suction cups are usually adopted for {surface attachment} due to their durability and large adhesive force. To seal on  {a surfaces}, suction cups {must} be aligned with {the surface} and {possess proper relative tangential velocity}. {However, quadrotors' attitude and relative velocity errors would become significant when the object surface is moving and inclined. To address this problem, we proposed a real-time trajectory planning algorithm. The time-optimal aggressive trajectory is efficiently generated through multimodal search in a dynamic time-domain. The velocity errors relative to the moving surface are alleviated.} To further adapt to the residual errors, we design a compliant gripper using self-sealing cups. Multiple cups in different directions are integrated into a wheel-like mechanism to increase the tolerance to attitude errors. The wheel mechanism also eliminates the requirement of matching the attitude and tangential velocity. {Extensive tests are conducted to perch on static and moving surfaces at various inclinations.} Results demonstrate that our proposed system enables a quadrotor to reliably perch on moving inclined surfaces (up to $1.07m/s$ and $90^\circ$) with a success rate of $70\%$ or higher. {The efficacy of the trajectory planner is also validated. Our gripper has larger adaptability to attitude errors and tangential velocities than conventional suction cup grippers.} The success rate increases by 45\% in dynamic  perches.
\end{abstract}

\def\abstractname{Note to Practitioners}
\begin{abstract}
This paper was motivated by the problem of perching on moving inclined surfaces using  quadrotors. It can be used for perching on the various angled surfaces of an automobile to save energy and enlarge flight distance. {It can also be exploited in air-ground cooperative tasks.} In recent years, various grippers and trajectory planning methods have been devised to enable quadrotors to perch on static inclined surfaces. These strategies can not be used for perching on moving inclined surfaces. The task poses high requirements for the efficiency of trajectory planning  and the grippers' adaptability to pose errors. This paper proposes a real-time planning method to generate trajectories. The trajectories respect kinematic and motor lift constraints. They allow large attitude maneuvering of quadrotors. A multimodal search in a dynamic time-domain is developed to seek the minimum feasible time for the trajectory planner. Considering large pose errors in dynamic perching, a compliant gripper based on multiple independent suction cups is designed. The wheel and multidirectional layout of cups increase the adaptability to attitude and velocity errors. Results suggest that the proposed system is effective and reliable. In practice, our multiple cups would add weight penalty and increase the drag. The suction cups should be placed close to a quadrotor's central axis to reduce the drag and disturbance in flow. The proposed system relies on an external motion capture system. In future research, we will focus on onboard sensing and control methods to achieve perching on moving inclined surfaces outdoors.
\end{abstract}


%
\IEEEpeerreviewmaketitle

\section{Introduction}

\IEEEPARstart{M}{icro} Aerial Vehicles (MAVs), specifically quadrotors, suffer from limited battery power, which significantly restricts their operation time and range \cite{hang2019perching}. {Perching on a surface could be helpful in energy conservation  and extending the operation time \cite{yu2020perching,yu2019exploring}.} {Therefore, recent years have seen a growing interest of the research community towards the perching of MAVs \cite{pope2016multimodal,roderick2017touchdown,kovac2016learning}.} However, the existing works are confined to perching on static objects. That could lengthen the operation time but not the range of motion. {In reality, many objects like vehicles are moving. The surfaces on the vehicles are not always horizontal. Some of them are inclined with different angles.} Thus, the set of possible perch locations would be enlarged if quadrotors are capable of perching on these surfaces. Furthermore, MAVs could mimic the hitchhiking behavior of remoras \cite{wang2017biorobotic} to increase the operational range. {MAVs could also perch on the} transport system to increase the efficiency of package delivery and provide convoy support \cite{paris2020control,lasla2019exploiting,farrell2020error,choudhury2021efficient,baca2019autonomous}. This mode can also be exploited in cooperative air-ground reconnaissance missions \cite{8798870}. {For example, quadrotors can charge and exchange information with the help of mobile platforms.} Therefore, the ability to perch on moving surfaces with various inclinations is promising in a wide range of missions. It should be thereby studied and improved sufficiently.

\subsection{Related Work}
To perch on moving inclined surfaces, two essential aspects should be considered: gripper design and trajectory generation. In terms of gripper design, electrostatic grippers are commonly adopted in quite small drones due to limited attraction force \cite{graule2016perching,chirarattananon2016perching}. Magnets are employed to perch on an inclined surface in \cite{huang2021biomimetic,zhang2017spidermav}. The method is restricted to ferromagnetic surfaces. The authors in \cite{kalantari2015autonomous} and \cite{thomas2016aggressive} use dry adhesives-based gripper for vertical or inclined surfaces perching. They are limited to clean and air-purified environments because dust and other contaminants can significantly decrease adhesion and lead to failure \cite{alizadehyazdi2020optimizing}. In comparison to dry adhesives, suction cups could work well repeatedly on unclean non-porous surfaces \cite{aoyagi2020bellows}. The adhesion force is also relatively large \cite{tsukagoshi2021soft}. {These attributes motivate our choice of} suction cups as grippers.  

One challenge in employing a suction cup gripper is to seal the cup and surface. To achieve sealing in the presence of the quadrotor's attitude errors, Tsukagoshi et al.\cite{tsukagoshi2015aerial} use a flexible tube as the support of the suction cups to increase the tolerance for angular errors. {The authors in \cite{huang2022enabling} employ a flexible universal joint to accommodate different shapes of the surface. They do not consider the adverse effect of tangential velocity. At the time of surface contact, tangential velocities of the cup relative to the surface will cause friction. The friction may bend the suction cup and make it difficult to close the cups and surfaces.} To make sealing easy while gripping objects of different sizes, Kessens et al.\cite{kessens2010design} employ multiple self-sealing suction cups in the same direction and achieve versatile aerial grasping \cite{kessens2016versatile} and perching \cite{kessens2019toward}. This co-directional arrangement cannot increase the adaptability to angle errors and tangential velocities. {As for the layout of the gripper, previous researchers in \cite{wopereis2016mechanism,du2015unified,huang2019design,liu2014impedance,kessens2019toward} mount suction cups on the sides of quadrotors. They achieve perching on inclined or vertical surfaces.} This layout requires an extra moment to compensate for the center of gravity excursion. It decreases the agility and controllability of MAVs. The layout also renders the grippers unlikely to be helpful for some potential tasks such as object transportation \cite{thomas2016aggressive,saunders2021autonomous}. {If grippers are mounted on the bottom of quadrotors, the large attitude maneuvering is necessary to perch on inclined surfaces. In this case, the angle and tangential velocities errors will be significant, especially in moving surface perching applications.
Therefore, a suction cup gripper's adaptability to angle errors and tangential velocities is essential. Further research is needed in this area.}  

As for trajectory generation, the trajectory planner is first required to be run in real-time to adapt to the movement of surfaces. {Second, the trajectories should allow the quadrotors to fly aggressively respecting kinodynamic constraints.} In this manner, the attitude of quadrotors can be exploited to align with surfaces at various inclinations. Third, the flying time should be as short as possible to reduce the change of surface states during the perching process. In existing work, {Wang et al.\cite{wang2018quadrotor} develop an on-board monocular vision system and achieve autonomous quadrotor landing on moving ship decks. The authors in\cite{gautam2018log} design a log polynomial velocity profile to implement autonomous landing on a moving platform. These two studies only consider planning position or velocity profile in altitude and landing on slightly inclined surfaces.} {Thomas et al.\cite{thomas2016aggressive} take the accelerations as the states to be planned. They develop a minimum snap trajectory generation method to enable a quadrotor to fly aggressively to perch on vertical surfaces.} The trajectory is planned offline and the shortest terminal time is also not considered. Thus, it can not be used for perching on moving surfaces. The authors in \cite{baca2019autonomous} simplify a quadrotor model into a third-order linear time invariant (LTI) system. A reference trajectory is generated by model predictive control (MPC) in real-time to land a drone on a moving car. {The accelerations are treated as system inputs. They are conservative and make the drone unable to achieve a large attitude angle.} Time optimization is also not considered. The authors in \cite{hu2019time} select thrust and angle rates as input. A time-optimal trajectory is planned to land a quadrotor onto a slightly heaving and tilting platform. In \cite{9669934}, velocity and attitude angle are treated as control inputs to tracking mobile ground vehicles. {In these works, trajectories are discretized as static decision variables. MPC problems are formulated and then solved by nonlinear programming.} This can incur high computational {costs or failure of the solver to} converge for large attitude maneuverability. A search-based planning method to generate feasible and time-optimal trajectories is proposed in \cite{liu2017search} and improved by \cite{zhou2019robust} to achieve fast autonomous flight in unknown environments. They also employ accelerations as inputs and large attitude maneuverability can not be planned. {
Mao et al.\cite{mao2021aggressive} adopt polynomial splines to represent trajectories. Trajectories are obtained by iteratively solving a quadratic programming problem. Perching on a static vertical surface is achieved. The planning algorithm considers thrust constraints.} Above all, in some extreme cases,  the feasibility of acceleration or angle rate does not guarantee the feasibility of motor lift. {This may cause infeasibility of large attitude maneuvering. In brief, the real-time planning method to generate time-optimal trajectories respecting constraints on motor lift is still an open research topic.} 

\subsection{Contribution of This Article}
With the aforementioned consideration, we devised  a compliant gripper using self-sealing suction cups built upon our previous work \cite{liu2020adaptive}. {The cups are integrated into a passive wheel in multiple directions. Thus, larger attitude errors can  be adapted.} The wheel mechanism in the gripper also eliminates the adverse coupling effect between attitude errors and tangential velocity. Therefore, the adaptability to tangential velocity is increased. For the second problem, we develop a method to generate trajectories in real-time. The differential flatness is exploited to check the feasibility of the quadrotor's motor lift. A multimodal search method in a dynamic time-domain is developed to obtain optimal feasible time. Time-optimal trajectories that satisfy the dynamics constraints can be efficiently generated. Therefore, aggressively perching on moving inclined (even vertical) surfaces is reliably achieved. To our knowledge, this is the first work performing  perching on a moving object.

The main contributions of this paper are twofold: (1) We design and analyze a gripper with large adaptability to pose errors and tangential velocities. It adopts multiple independent self-sealing suction cups. (2) We demonstrate a method to plan time-optimal aggressive trajectories in real-time and control a quadrotor to perch on a moving inclined surface.

The remainder of this paper is organized as follows: The next section presents the design of the seal-sealing suction cup, gripper and quadrotor system. Section \uppercase\expandafter{\romannumeral3} contains the real-time planning and control method to perch on a moving surface. Section \uppercase\expandafter{\romannumeral4} describes the tests of negative pressure and experiments of perching on static and moving surfaces with different inclinations. The corresponding results and discussion are also presented. Finally, conclusions and future work are given in Section \uppercase\expandafter{\romannumeral5}.

\section{Description of The Perching System }
We describe the designed system in three subsections. In subsection  \uppercase\expandafter{\romannumeral2}-A, the overview and key dimensions of a self-sealing cup are presented. The mechanism of an inner trigger cup is detailed. The corresponding working process of the self-sealing cup module is analyzed. In subsection \uppercase\expandafter{\romannumeral2}-B, the gripper based on self-sealing cups is described. In subsection \uppercase\expandafter{\romannumeral2}-C, the whole aerial perching system is demonstrated.

\subsection{Self-sealing Suction Cup}
\begin{figure}[!tb]
\centering
\includegraphics[width=3.0in]{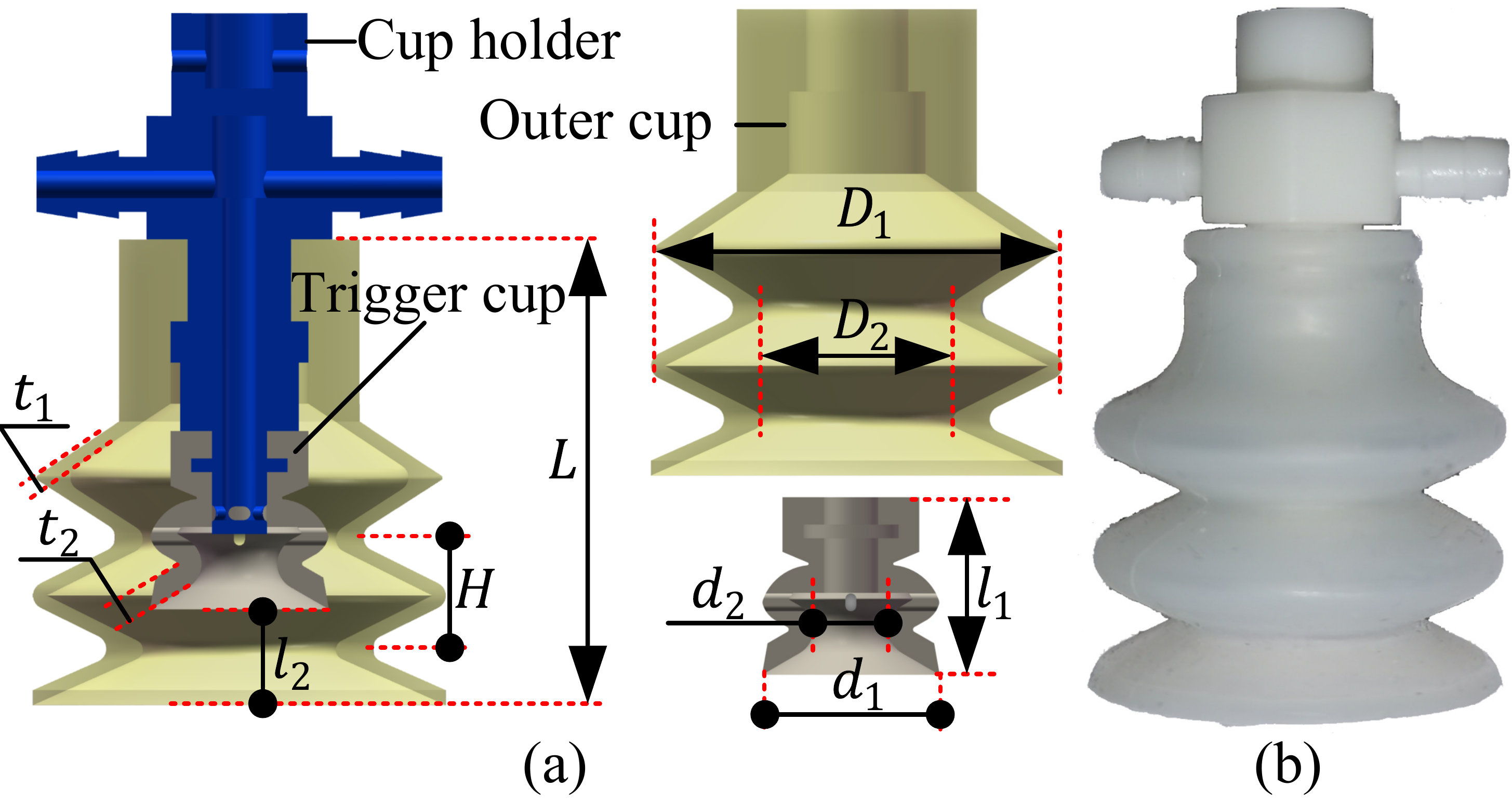}
\caption{The overview of the self-sealing suction cup. (a) The dimension annotations. (b) The real product of the self-sealing suction cup. }
\label{cupmodule}
\end{figure}

The self-sealing suction cup (SS-Cup) is detailed in Fig.\ref{cupmodule}. The SS-Cup contains an inner trigger (gray part) and an outer cup (yellow part), a rigid cup holder (blue part). The main dimensions are labeled in Fig.\ref{cupmodule} (a) and the specific parameters are {shown} in Table. \ref{table_dimensionsofcup}. The outer cup is made of silicone rubber (45 Shore A). The trigger cup is made of Hei-Cast 8400 (50 Shore A). The cup holder is made of VeroWhitePlus photosensitive resin by 3D printing. A real product of the SS-Cup is shown in Fig.\ref{cupmodule} (b).

\begin{table}[!t]
  \centering
  \renewcommand{\arraystretch}{1.3}
  \caption{MAIN DIMENSION PARAMETERS OF SS-Cup ({\upshape mm})}
  \label{table_dimensionsofcup}
  \begin{tabular}{cccccccccc}
    \toprule[1.2pt]
   $D_1$ & $D_2$ & $d_1$ & $d_2$ & $L$ & $l_1$ & $l_2$ & $t_1$ & $t_2$ & $H$\\
    \midrule[1pt]
    30	 & 14	& 13 &	5.5 &	34 &	13 &	7 &	1 &	1.8 &	8\\
    \bottomrule[1.2pt]
  \end{tabular}
  \end{table}
\begin{figure}[!tb]
\centering
\includegraphics[width=3.0in]{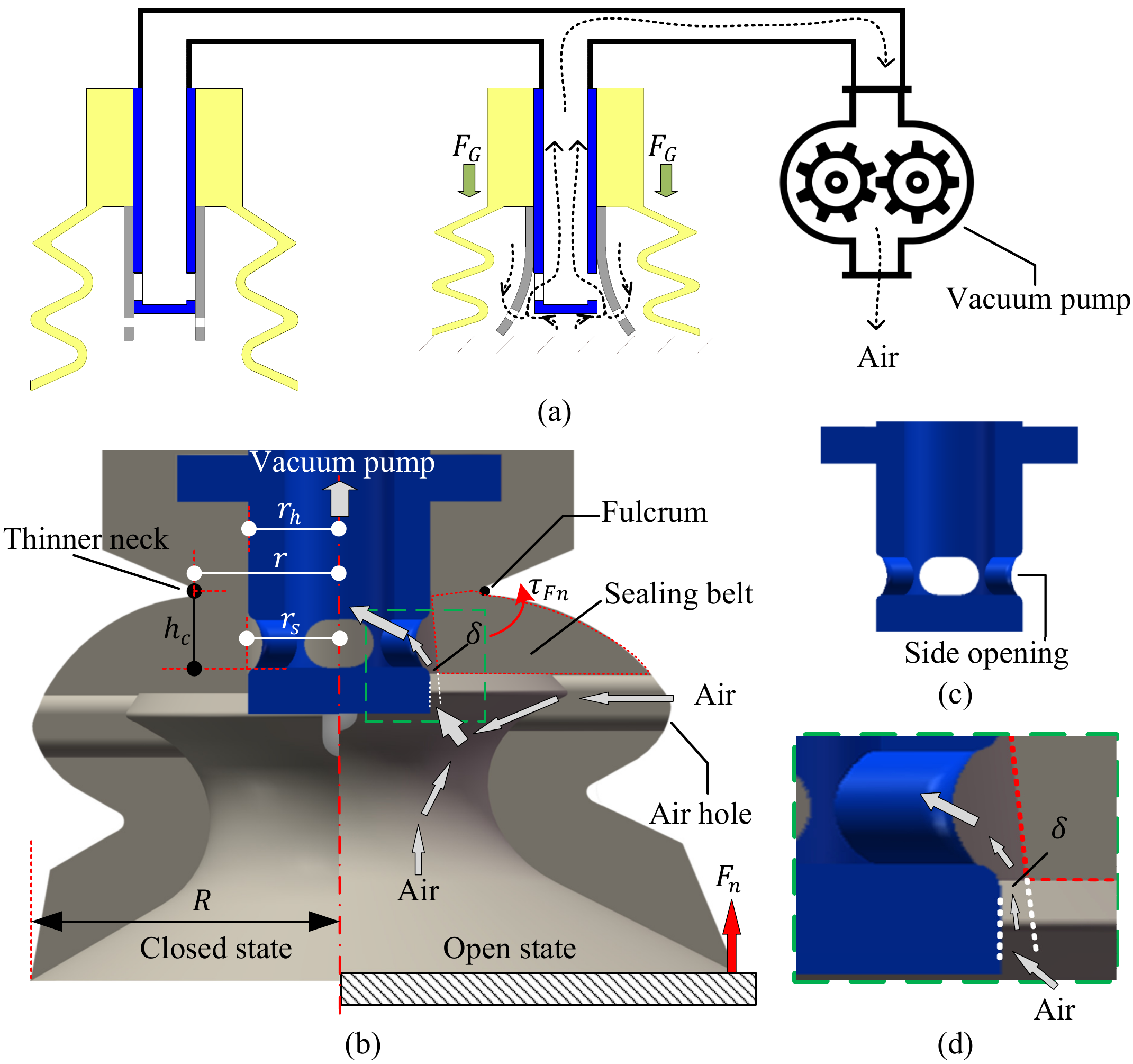}
\caption{{The working logic of multiple SS-cups and mechanism of the self-sealing structure in trigger cups. (a)The diagram of two SS-cups connect one vacuum pump. (b) The closed state and open state of the inner trigger cup. (c) The side opening in the cup holder. (d) The close-up view of the gap $\delta$  in the open state.}}
\label{innercupdeformation}
\end{figure}
{The working logic of multiple SS-cups sharing one vacuum pump is shown in Fig.\ref{innercupdeformation} (a). The object surface only contacts the right SS-cup but not the left one. To engage with the surface, the air in the right SS-cup should be evacuated. At the same time, no air is allowed to enter the system through the left SS-cup. In the diagram, the gray part is the inner trigger cup. The blue part is the cup holder with side openings (shown in Fig.\ref{innercupdeformation} (c)). For the left SS-cup, the inner trigger cup compresses the side opening tightly so that no air leaks through the left SS-cup. The state of the left trigger cup is called the closed state. As for the right SS-cup, the trigger cup is deformed by the force from the object surface. The side opening is exposed due to the outward deformation. Then, the air in the SS-cup is evacuated through the side opening. The state of the right trigger cup is therefore called the open state.  This way, the trigger cup can be self-closed when the SS-cup is not employed. It  can also be self-opened when the SS-cup is pressed on a surface. This function is realized by our designed self-sealing structure in trigger cups.}

{The details of a self-sealing structure in a trigger cup are shown in Fig.\ref{innercupdeformation}(b). The left half part shows the closed state while the right half part shows the open state. $F_n$ is the normal reaction force from the object surface. The force will generate a torque $\tau_{F_n}$ around the thinner neck. The thinner neck can be regarded as a fulcrum. Other parts, especially the sealing belt, can rotate around the fulcrum under torque $\tau_{F_n}$. Then, a gap $\delta$ will emerge between the side opening and the sealing belt. If $\delta$ is large enough, the air in the trigger cup can be evacuated via $\delta$ and the side opening. Then, the trigger cup enters the open state. The details of $\delta$ in the green dotted box are shown in Fig.\ref{innercupdeformation} (d).}

In our previous work \cite{liu2020adaptive}, we formulate the relationship between $F_n$ and the gap $\delta$. Herein, we directly give the result as (\ref{eqn_activeforce}):
\begin{equation}
  F_n\approx\frac{(\delta+r_h-r_s)k}{(R-r)h_c}. \label{eqn_activeforce}
\end{equation}

{Where $r_h$ is the radius of the cup holder at the side opening. $r_s$ is the radius of the sealing belt. $k$ is the equivalent stiffness coefficient. $R$ is the radius of the trigger cup. $r$ is the radius of the thinner neck. $h_c$ is the vertical distance of the fulcrum and the lower edge of the side opening.} In (\ref{eqn_activeforce}), the gap $\delta$ will increase with the increasing force $F_n$. When $\delta$ exceeds a certain threshold, {the trigger cup is regarded as open. The minimum force to turn the trigger cup into the open state is termed activation force. According to (\ref{eqn_activeforce}), the activation force can be modulated by adjusting $r_h$. Compared with the design in \cite{kessens2010design}, the proposed self-sealing structure is easier to manufacture and regulate.}

\begin{figure}[!tb]
  \centering
  \includegraphics[width=3.0in]{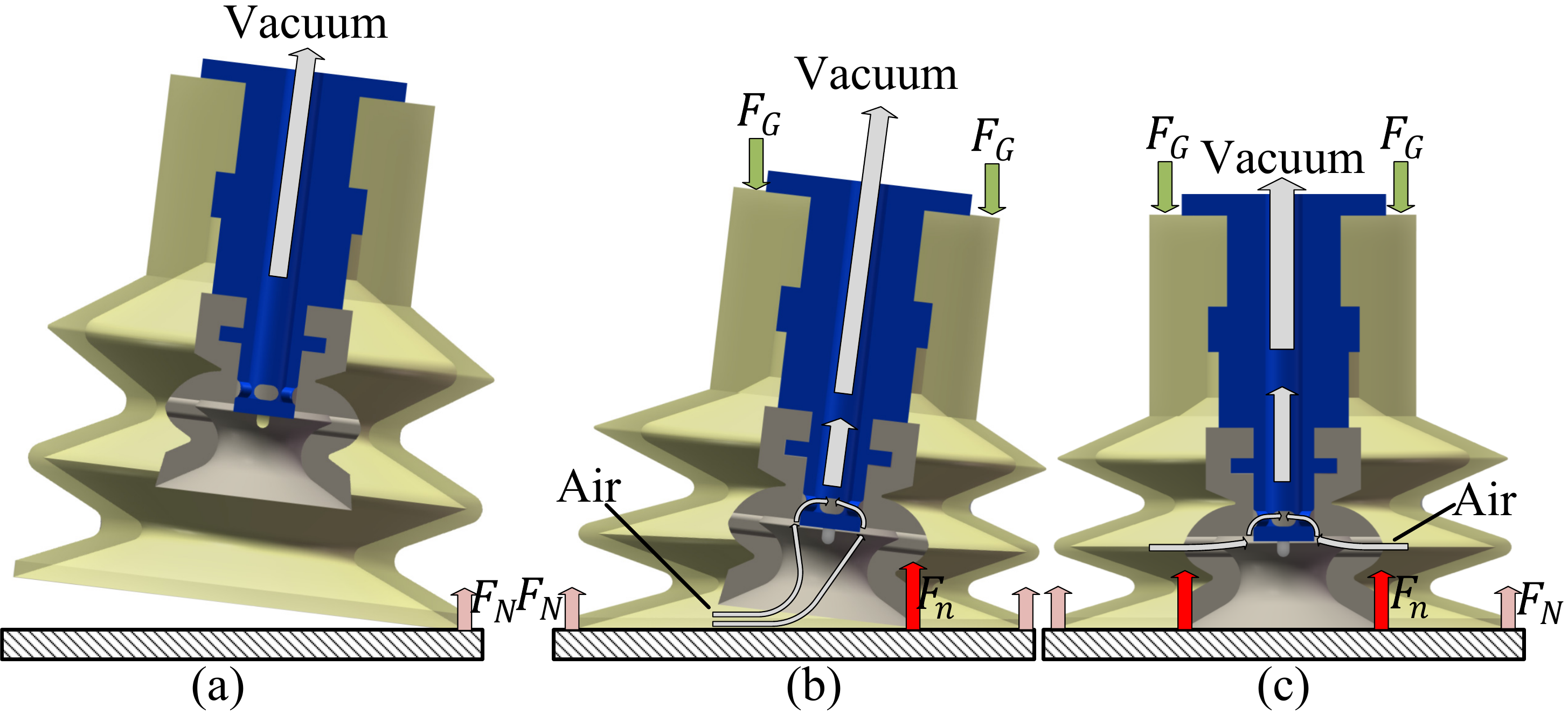}
  \caption{{The working principle of a SS-Cup. (a) Initial contact of the SS-Cup. (b) Contact force $F_N$ seals the SS-cup to a surface. The trigger cup is opened by the surface.(c) The air between the trigger cup and the outer cup is further evacuated through air holes on the trigger cup.}
  \label{cupmoduleworkingprocess}}
  \end{figure}
The working principle of a single SS-cup is shown in Fig.\ref{cupmoduleworkingprocess}. In this process, we assume the SS-Cup has no tangential velocity and moves perpendicular to a surface. Initially, the outer cup contacts a surface obliquely (Fig.\ref{cupmoduleworkingprocess} (a)). Then, the seal is formed owing to the outer cup's deformation caused by the reaction force $F_N$. The trigger cup may be opened by the reaction force $F_n$. Once the trigger cup is opened, the air in the SS-Cup will be evacuated (see Fig.\ref{cupmoduleworkingprocess} (b)). Particularly, the air between the outer cup and the trigger cup can be evacuated through the air holes on the trigger cup (Fig.\ref{cupmoduleworkingprocess} (c)). {Then, the force $F_G$ induced by the pressure difference will generate (see Fig.\ref{cupmoduleworkingprocess} (b)). The suction cup will be further compressed on the surface by $F_G$. This will increase $F_n$ to open the self-sealing structure more sufficiently. Then, the air can be further evacuated. That is, positive feedback can be formed once the suction cup is activated. Finally, this makes the negative pressure equal the pump's maximum vacuum degree.}

\subsection{Multi-directions Gripper Based on SS-Cup}
\begin{figure}[!tb]
  \centering
  \includegraphics[width=3.0in]{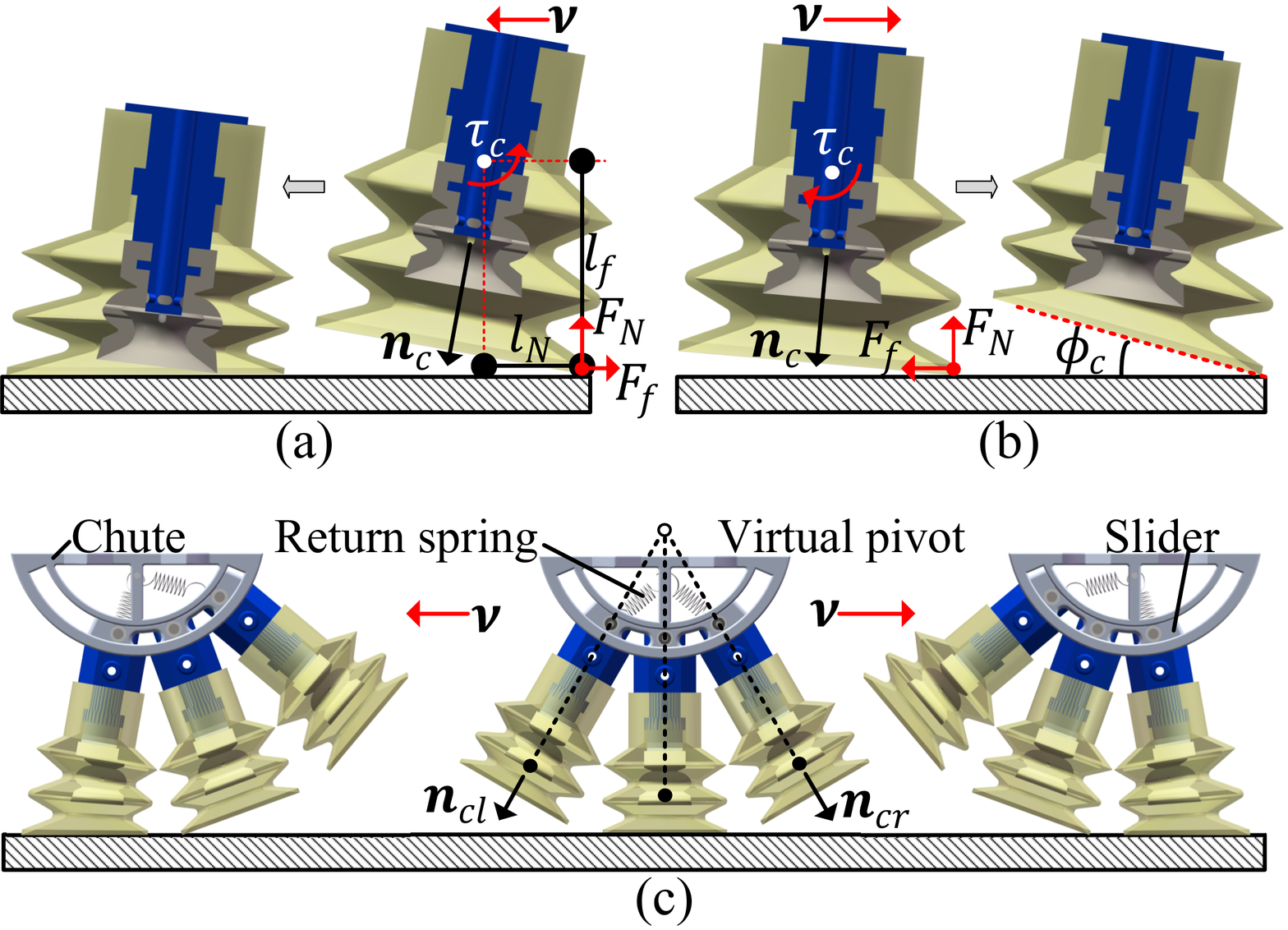}
  \caption{The gripper based on SS-Cup to accommodate tangential velocity in different directions. (a) The angular errors could be eliminated if $\boldsymbol{n}_c \cdot \boldsymbol{\nu} \geq 0$. (b) The angular errors might increase if $\boldsymbol{n}_c \cdot \boldsymbol{\nu} < 0$. (c) The gripper consisting of three SS-Cups in multi-directions could accommodate tangential velocity in different directions.}
  \label{velocityeffectandwheeldesign}
  \end{figure}
We assume the SS-Cup has no tangential velocity when we analyze its working process in subsection \uppercase\expandafter{\romannumeral2}-A. Here, we will consider the influence of the tangential velocity on the SS-Cup. Note that the analysis is also suitable for conventional suction cups. {We define the cup direction as $\boldsymbol{n}_c$. It is perpendicular to the end of the cup and facing outward (Fig.\ref{velocityeffectandwheeldesign} (a,b)).} The cup's tangential velocity relative to the surface is $\boldsymbol{\nu}$. The center of the SS-Cup is denoted by the white dot. The torque $\tau_c$ acting on the center of the SS-Cup can be expressed as followings:
\begin{equation}
\tau_c=\left\{
\begin{aligned}
F_Nl_N+F_fl_f & & {\boldsymbol{n}_c \cdot \boldsymbol{\nu} \geq 0}\\
F_Nl_N-F_fl_f & & {\boldsymbol{n}_c \cdot \boldsymbol{\nu} < 0}.
\end{aligned}
\right.
\label{eqn_torquevelocity}
\end{equation}

As is shown in  (\ref{eqn_torquevelocity}), if $\boldsymbol{n}_c \cdot \boldsymbol{\nu} \geq 0$, $\tau_c$ is positive and could reduce the angular errors (Fig.\ref{velocityeffectandwheeldesign} (a)). On the contrary, if $\boldsymbol{n}_c \cdot \boldsymbol{\nu} < 0$, {$\tau_c$ could be negative and enlarge the angular errors $\phi_c$(Fig.\ref{velocityeffectandwheeldesign} (b)).} Therefore, the attitude of the single cup must be matched with the tangential velocity. To accommodate $\boldsymbol{\nu}$ in different directions, we design a gripper comprising of three SS-Cups arranged in multi-directions (MD-Gripper) (see Fig.\ref{velocityeffectandwheeldesign} (c)). {These cups are mounted on a slider which can slide in an arc chute. It is equivalent to rotating around a virtual pivot. If the chute moves to the left, the slide will rotate counterclockwise due to the friction $F_f$.  Then, the left SS-cup could be used and $\boldsymbol{n}_{cl} \cdot \boldsymbol{\nu} \geq 0$ is satisfied (left part of Fig.\ref{velocityeffectandwheeldesign} (c)).} On the contrary, the right cup will be used and $\boldsymbol{n}_{cr} \cdot \boldsymbol{\nu} \geq 0$ {is also satisfied} (right part of Fig.\ref{velocityeffectandwheeldesign} (c)). Therefore, the MD-Gripper could always satisfy $\boldsymbol{n}_{c} \cdot \boldsymbol{\nu} \geq 0$ for $\boldsymbol{\nu}$ in different directions. The proposed MD-Gripper eliminate the matching requirement of the attitude and tangential velocity. Additionally, the slider-chute mechanism could also keep the chute and the surface parallel when cups roll left or right. {This could prevent the quadrotor (see Fig.\ref{quadrotorsystem}) from colliding with the surface. Two springs can return the slider to the middle position when there is no tangential force.}
\subsection{Aerial Perching System}
\begin{figure}[!tb]
\centering
\includegraphics[width=2.4in]{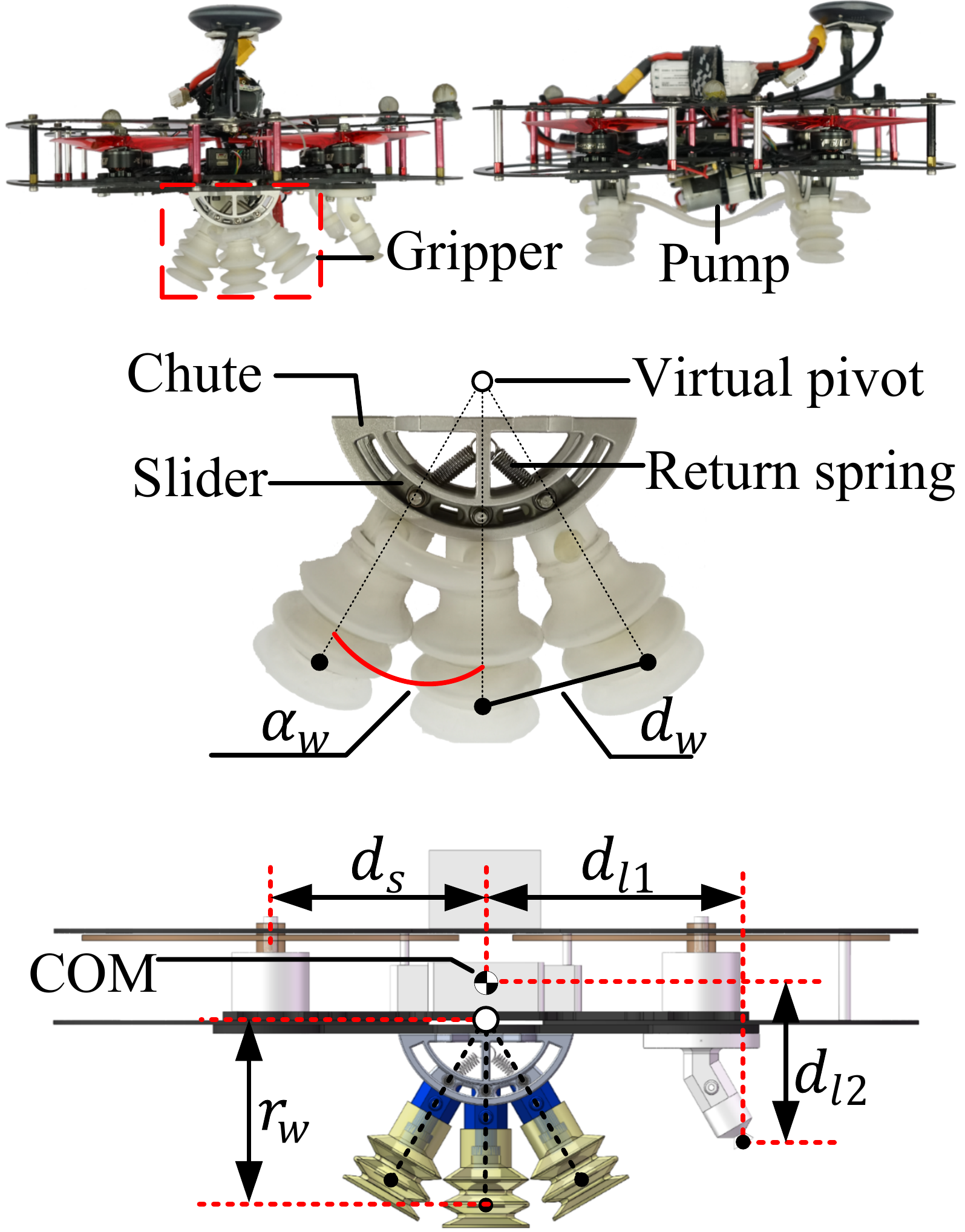}
\caption{The Aerial Perching System. The main dimensions of the aerial perching system and the real product of the system and gripper.}
\label{quadrotorsystem}
\end{figure}
\begin{table}[!t]
  \centering
  \renewcommand{\arraystretch}{1.0}
  \caption{MAIN  PARAMETERS OF AN AERIAL PERCHING SYSTEM}
  \label{table_dimensionsofquad}
  \setlength{\tabcolsep}{1.5mm}{
  \begin{tabular}{cccc|c|ccc}
    \toprule[1.2pt]
      \multicolumn{4}{c|}{Length ({\upshape mm})} & Angle ($^{\circ}$)& \multicolumn{3}{|c}{Mass ({\upshape kg})}
      \\ \midrule[1pt]
    $r_w$    &  $d_s$  & $d_{l1}$   & $d_{l2}$  & $\alpha_w$  & Gripper & Pump& System  \\ \midrule[1pt]
     65.7   &   79.2   &  85.9  & 53.9   &    30   &  0.052       &  0.053      & 0.945    \\
    \bottomrule[1.2pt]
  \end{tabular}}
 \end{table}
The real product of a perching system and  main dimension annotations are presented in Fig.\ref{quadrotorsystem}. The system contains a quadrotor based on carbon board fuselage and PX4-PixracerR15 autopilot, a vacuum pump cup and two grippers. {The suction cups in the grippers connect with the pump directly without any valves (see Fig.\ref{innercupdeformation} (a)). The pump is directly connected to the power supply. It will run all the time once the system is powered on.} The virtual pivot is the rotation center of the slider. Then, the three cups could roll like a wheel. The radius of the wheel is denoted as $r_w$. The chord length between the endpoints of two trigger cups is denoted as $d_w$. The angle between two adjacent SS-Cups is $\alpha_w$. The center of mass (COM) is approximately the geometric center of the quadrotor. $d_{l1}, d_{l2}$ are series distances to COM (see Fig.\ref{quadrotorsystem}). The specific parameters are {shown} in Table. \ref{table_dimensionsofquad}. It should be noted that the grippers and pump add the system weight by $11.1\%$. The wheel mechanism also  increases the air resistance area. This will increase the drag and disturb the flow from rotors. To alleviate these effects, the gripper should be placed near the center of the quadrotor. 

\section{Planning and Control}
In this section, we will explain our real-time trajectory planning and tracking method. This section contains three parts. {In subsection III-A, we briefly introduce the dynamics of the drone in a sagittal plane. The minimum jerk trajectory planning method is also presented. In subsection III-B, we demonstrate an algorithm to search the optimal terminal time for the minimum jerk trajectory planner.} In subsection III-C, the trajectory tracking method and the way to determine terminal states are demonstrated.
\subsection{Dynamics and Minimum Jerk Trajectory Generation}
\begin{figure}[!tb]
  \centering
  \includegraphics[width=3.0in]{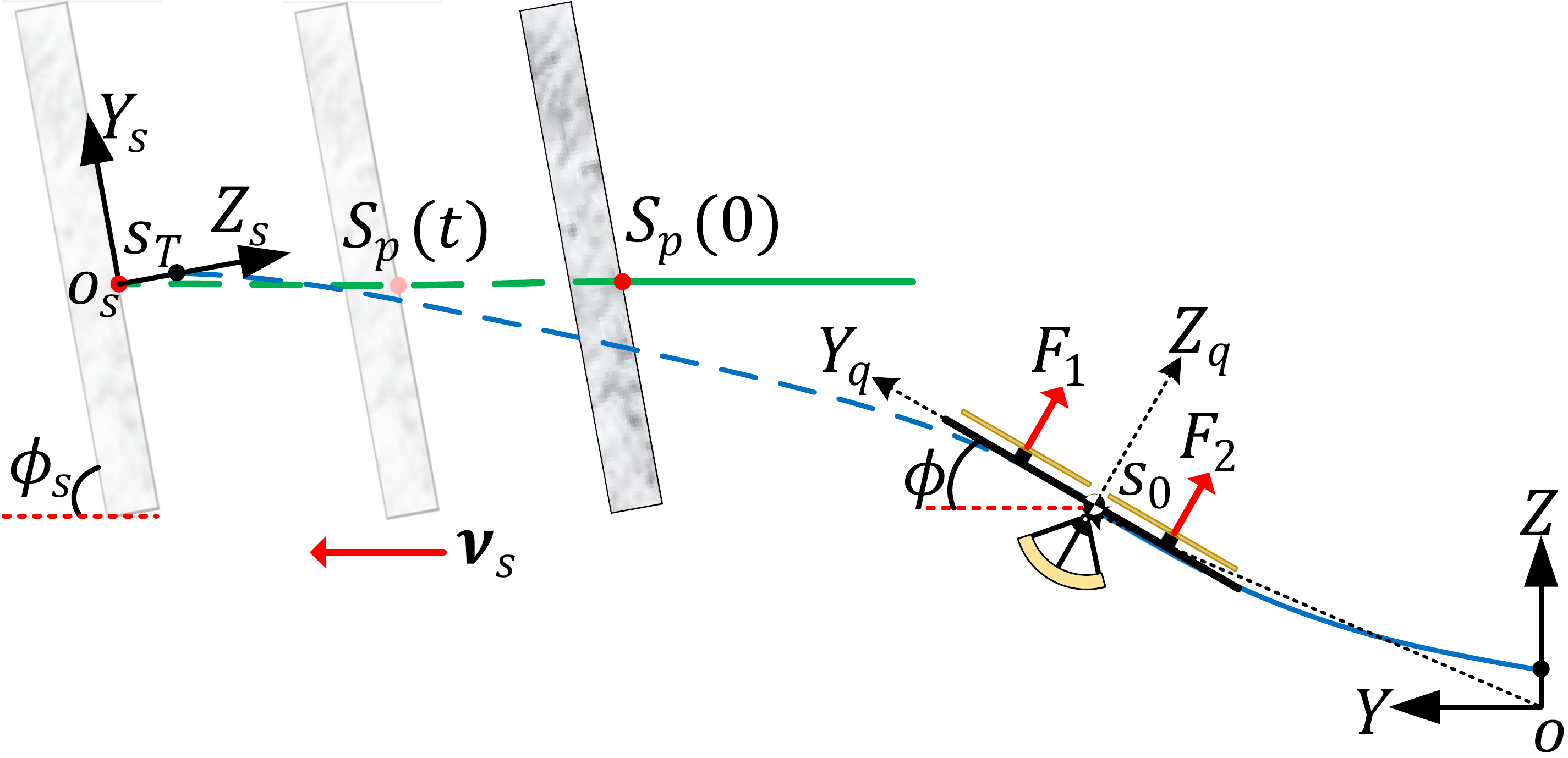}
  \caption{The coordinate system in perching task. {The blue dash line is the planned trajectory. The green dash line is the predicted trajectory of the surface. The green solid line is the history trajectory.}}
  \label{coordinatemodelforplancontrol}
  \end{figure}
Generally, a vehicle moves in its symmetry plane and the direction changes relatively slowly. Therefore, it is natural to analyze the problem in the symmetry sagittal  plane. The coordinate system is illustrated in Fig.\ref{coordinatemodelforplancontrol}. There are three coordinates in our setting: the static global coordinate ($Y-O-Z$), the quadrotor body coordinate ($Y_q-O_q-Z_q$) and the surface coordinate ($Y_s-O_s-Z_s$). {All vectors in this article are represented in the global coordinate system unless otherwise specified.} The quadrotor is {controlled}  by the lift of two pairs of propellers $F_1$ and $F_2$. {The planar model has three degrees of freedom: the horizontal and vertical position $y$ and $z$, and the roll angle $\phi$.} The equations of motion in symmetry sagittal  plane are \cite{hu2019time,hehn2012performance}:

\begin{footnotesize}
\begin{equation}
  \begin{split}
  \ddot y\! =\!-\dfrac{F_1+F_2}{m}\sin{\phi},\;
  \ddot z \!=\!\dfrac{F_1+F_2}{m}\cos{\phi}-g,\;
  \ddot{\phi} \!=\!\dfrac{(F_1-F_2)d_s}{J}.
  \end{split} \label{eqn_dynamics}
\end{equation}
\end{footnotesize}
Where $J$ denotes the moment of inertia with respect to the $X_q$ axis of the quadrotor. $d_s$ is defined in Fig.\ref{quadrotorsystem}. This system is differentially flat with the coordinate $\boldsymbol{r}_q=(y,z)$ as flat output and input $\boldsymbol{u}=(F_1,F_2)$. According to equations in (\ref{eqn_dynamics}), we can get
\begin{gather}
  \phi=-\arctan{\dfrac{\ddot y}{{\ddot z}+g}} \label{eqn_phi} \\
   F_1+F_2=m\sqrt{{({\ddot z}+g)}^2+{\ddot y}^2} \label{eqn_f1+f2} \\
   F_1-F_2=\dfrac{\ddot{\phi}J}{d_s}=-\left(\frac{{\dddot y}({\ddot z}+g)-{\ddot y}{\dddot z}}{{({\ddot z}+g)}^2+{\ddot y}^2} \right)^{'} \cdot \dfrac{J}{d_s}. \label{eqn_f1-f2}
\end{gather}
From (\ref{eqn_f1+f2})-(\ref{eqn_f1-f2}), we can solve the input $F_1$ and $F_2$. They are the function of the flat outputs $y$ and $z$ and their derivatives. This property makes the planning of flat output states sufficient to represent the full states of the system.

In trajectory generation, each component of $\boldsymbol{r}_q$ is planned independently. Here, we take the component $y$ as an example to illustrate the trajectory generation. The method is based on minimum jerk criteria, and it is described as
\begin{equation}
  \begin{split}
  \min \quad &J =\int_{0}^{T}{\frac{\mathrm{d}^{3} y }{\mathrm{d} t^{3}}}^2dt\\ 
  s.t. \quad &y(0)=y_0, \; {\dot y}(0)={\dot y}_0, \; {\ddot y}(0)={\ddot y}_0\\
        \quad &y(T)=y_T, \; {\dot y}(T)={\dot y}_T, \; {\ddot y}(T)={\ddot y}_T.
  \end{split} \label{eqn_minimumjerk}
\end{equation}
The optimal solution of (\ref{eqn_minimumjerk}) can be obtained using Pontryagin's minimum principle as follows\cite{dong2015ball}:
\begin{equation}
\setlength{\arraycolsep}{1.0pt}
\footnotesize{
\left[
\begin{array}{c}
    y \\ 
    \dot y \\
    \ddot y \\
    \dddot y
\end{array}
\right]=
\left[
\begin{array}{c}
    \dfrac{c_1}{120}t^5+\dfrac{c_2}{24}t^4+\dfrac{c_3}{6}t^3+\dfrac{{\ddot y}(0)}{2}t^2+{\dot y}(0)t+y(0) \\ 
    \dfrac{c_1}{24}t^4+\dfrac{c_2}{6}t^3+\dfrac{c_3}{2}t^2+{\ddot y}(0)t+{\dot y}(0) \\
   \dfrac{c_1}{6}t^3+\dfrac{c_2}{2}t^2+{c_3}t+{\ddot y}(0)\\
    \dfrac{c_1}{2}t^2+{c_2}t+c_3 
\end{array}
\right]}
\label{eqn_formintrjsolution}
\end{equation}
with
\begin{equation}
\setlength{\arraycolsep}{1.0pt}
\left[
\begin{array}{c}
    c_1 \\ 
    c_2 \\
    c_3 
\end{array}
\right]=\dfrac{1}{T^5}
\left[
\begin{array}{ccc}
    60T^2 & -360T & 720\\
    -24T^3 & 168T^2 & -360T\\
    3T^4 & -24T^3 & 60T^2
\end{array}
\right]
\left[
\begin{array}{c}
    \Delta {\ddot y} \\ 
    \Delta {\dot y} \\
    \Delta y 
\end{array}
\right]
\label{eqn_formimtrjcoefficent}
\end{equation}

\begin{equation}
\left[
\begin{array}{c}
    \Delta {\ddot y} \\ 
    \Delta {\dot y} \\
    \Delta y 
\end{array}
\right]=
\left[
\begin{array}{c}
    {\ddot y}_T-{\ddot y}_0 \\ 
    {\dot y}_T-{\dot y}_0-{{\ddot y}_0}T \\
   y_T-y_0-{{\dot y}_0}T-\dfrac{1}{2}{{\ddot y}_0}T^2
\end{array}
\right].
\label{eqn_fordeltastates}
\end{equation}

From (\ref{eqn_formintrjsolution})-(\ref{eqn_fordeltastates}), the trajectory satisfying the initial and terminal states constraints in (\ref{eqn_minimumjerk}) can be solved given terminal time $T$. {The solutions for component $z$ follow similarly.}
\subsection{Minimum Time Search}
A minimum terminal time is preferred to reduce the state variation of the moving surface during perch. However, kinodynamic constraints need to be satisfied when we try to seek a shorter  terminal time. These constraints are formulated as:
\begin{gather}
  z(t) \in [z_{min},z_{max}] \quad \quad \dot y(t),\dot z(t) \in [v_{min},v_{max}] \label{eqn_conpos}\\
  F_1(t),F_2(t) \in [0,F_{max}].\label{eqn_conF1F2}
\end{gather}

Where the $z_{min},z_{max},v_{min},v_{max},F_{max} $ are the corresponding lower and upper boundaries. {In previous work, the constraints of the thrust and angle rate in \cite{van2013time} or the thrust and attitude angle in \cite{zhang2020attitude} are commonly adopted. In this article, we adopt the constraints of motor lifts ($F_1$ and $F_2$). The constraints guarantee deeper and more accurate dynamics compatibility.} After the trajectories are solved according to (\ref{eqn_formintrjsolution})-(\ref{eqn_fordeltastates}), we sample uniformly on the solved state and input profiles. {We will compare these sampling points to the constraints to check their feasibility.} If the trajectories violate constraints in (\ref{eqn_conpos})-(\ref{eqn_conF1F2}), the terminal time needs to be reselected. To find the minimum time, we provide an algorithm (Algorithm \ref{alg_timesearh}) based on {searching in a dynamic time-domain.}

\begin{algorithm}[!tb]
\caption{ Minimum Terminal Time Search }
\label{alg_timesearh}
\LinesNumbered 
\KwIn{
 $\boldsymbol{S}_0$;\quad $\boldsymbol{S}_p(t)$;\\
}
\KwOut{$T$;\quad $\boldsymbol{S}_T$;\\
}
$T_l=0.5T_{last}$; $T_r=1.5T_{last}$; $\delta T=(T_r-T_l)/5$;\\
FeasibleTestFlag=False;\\
\While{FeasibleTestFlag==False}{
    $\boldsymbol{S}_T$=GetTerminalStates($\boldsymbol{S}_p(T_l)$);\\
    FeasibleTestFlag=FeasibilityTestFunction($\boldsymbol{S}_0,\boldsymbol{S}_T,T_l$);\\
    \eIf{FeasibleTestFlag==False}{
        $T_l=T_l+\delta T$;\\
        \If{$T_l > T_r$}{
            $\delta T={\delta T}/2;\quad T_l=0.5T_{last}+\delta T$;\\
            \If{$\delta T < 0.01$}{
                break;
            }
        }
    }{
        $T_r=T_l; \quad T_l=T_r-\delta T$;\\
        \While{$T_r-T_l>0.1$}{
            $\boldsymbol{S}_T$=GetTerminalStates($\boldsymbol{S}_p((T_l+T_r)/2)$);\\
            \eIf{FeasibilityTestFunction($\boldsymbol{S}_0,\boldsymbol{S}_T,(T_l+T_r)/2$)==False}{
                $T_l=(T_l+T_r)/2$;
            }{
                $T_r=(T_l+T_r)/2$;
            }
        }
        break;
    }
}
\eIf{FeasibleTestFlag==True}{
    $T=T_r;\; T_{last}=T$; $T_e$=GetCurrentSystemTime();\\
}{
    $T=T_{last}$-(GetCurrentSystemTime()-$T_e$);\\
    $T_{last}=T$;\quad $T_e$=GetCurrentSystemTime();\\
}
\eIf{$T \geq 0.4$}{
    Return $T$ and GetTerminalStates($\boldsymbol{S}_p(T)$);\\
}{
    Return Null;
}
\end{algorithm}
There are two global variables used in Algorithm \ref{alg_timesearh}. $T_{last}$ is the minimum terminal time solved in the last loop. $T_e$ is the system time {at the end of the last solving loop.} In the first loop, we enumerate the terminal time from 0 and increasingly by a small step until a feasible time is searched. The searched feasible time and the time when it is searched are employed to initialize $T_{last}$ and  $T_e$. The input of the Algorithm \ref{alg_timesearh} is current states $\boldsymbol{S}_0=(y_0,{\dot y}_0,{\ddot y}_0,z_0,{\dot z}_0,{\ddot z}_0)$ of a quadrotor and the predicted trajectory $\boldsymbol{S}_p(t)=(y_s(t),{\dot y}_s(t),z_s(t),{\dot z}_s(t),{\phi}_s)$ of a moving surface. The output contains the minimum terminal time $T$ and the corresponding terminal states $\boldsymbol{S}_T=(y_T,{\dot y}_T,{\ddot y}_T,z_T,{\dot z}_T,{\ddot z}_T)$. Combing the input $\boldsymbol{S}_0$ and the output $T,\boldsymbol{S}_T$, the trajectories can be obtained according to (\ref{eqn_formintrjsolution})-(\ref{eqn_fordeltastates}). The surface is assumed to move along $Y$ axis with constant speed $\boldsymbol{\nu}_s$ and inclination angle $\phi_s$. {The first-order} least square method is used for fitting the historical trajectory and predicting the future position, velocity (see Fig.\ref{coordinatemodelforplancontrol}). $\boldsymbol{S}_p(0)$ denotes the current states of the surface and $\boldsymbol{S}_p(t)$ is the predicted states at $t$. 

Algorithm \ref{alg_timesearh} is executed whenever $\boldsymbol{S}_0$ or $\boldsymbol{S}_p(t)$ is updated. First, a search region is defined by left node $T_l$ and right node $T_r$. {The region $(T_l,T_r)$ centers on $T_{last}$ (Line 1-2). It is called dynamic time-domain due to $T_{last}$ constantly changing.} This is based on a natural assumption that the solved time corresponding to two adjacent states is near. Second, a feasible time is searched from $T_l$ to $T_r$ by a search step $\delta T$. If no feasible time was found, $\delta T$ will be refined. The search again starts from the initial left node  (Line 6-18). Once a feasible time is found, a dichotomizing search is adopted to find a minimum feasible {time} $T$ in the region $(T_l,T_r)$ (Line 19-29). For the situation where no feasible time is found, {$T$ is determined by subtracting an interval from $T_{last}$. The interval is the time from now to $T_e$ (Line 34-35).} This supplement scheme is necessary to ensure the terminal states are updated in time. As the quadrotor approaches the terminal states,  $T$ gets smaller. The solution will become unstable {because} $T^5$ is contained in the denominator in (\ref{eqn_formimtrjcoefficent}). Therefore, the planning program is stopped if $T$ is relatively small (Line 37-41). {During this period}, the last trajectories will be employed until the new trajectories are generated.

{The function \textit{FeasibilityTestFunction($\boldsymbol{S}_0,\boldsymbol{S}_T,T$)} is to verify the feasibility of $T$ according to (\ref{eqn_conpos})-(\ref{eqn_conF1F2}).} Another function \textit{GetTerminalStates($\boldsymbol{S}_p(t)$)} is to determine the terminal states of trajectories at $t$. {They are determined by $\boldsymbol{S}_p(t)$, conditions for successful perching and the control performance.} This part will be elaborated in subsection \uppercase\expandafter{\romannumeral2}-D.

\subsection{Trajectory Tracking Controller}
Although a 2D model is adopted in trajectory planning, {a 3D controller is necessary in reality.} This is formulated as:
\begin{equation}
  \begin{split}
  &\boldsymbol{\ddot r_q^c}=\boldsymbol{k}_p\boldsymbol{e}_p+\boldsymbol{k}_v\boldsymbol{e}_v+\boldsymbol{k}_i\int \boldsymbol{e}_pdt+\boldsymbol{k}_a \boldsymbol{\ddot r}_q^r\\
  &\boldsymbol{e}_p=\boldsymbol{r}_q^r-\boldsymbol{r}_q^a, \quad \boldsymbol{e}_v=\boldsymbol{\dot r}_q^r-\boldsymbol{\dot r}_q^a
  \end{split} \label{eqn_controller}
\end{equation}
\begin{equation}
\left\{
\begin{aligned}
&k_{ai}=\dfrac{1}{1+\dfrac{|k_{pi}e_{pi}+k_{vi}e_{vi}|}{|\ddot r_{qi}^r|+0.5}} & & {t \in [0, T-\delta t]}\\
&\boldsymbol{k}_{a}=\boldsymbol{I}, \; \boldsymbol{k}_p=\boldsymbol{0},\; \boldsymbol{k}_v=\boldsymbol{0},\;\boldsymbol{k}_i=\boldsymbol{0} & & {t \in (T-\delta t, T]}
\end{aligned}
\right.
\label{eqn_ka}
\end{equation}

Where $\boldsymbol{\ddot r_q^c}=(a_x^c,a_y^c,a_z^c)$ is the command acceleration containing three components. $\boldsymbol{r_q^r}=(x^r\equiv 0,y^r,z^r)$ is the reference trajectory. $y^r,z^r$ are obtained from (\ref{eqn_formintrjsolution})-(\ref{eqn_fordeltastates}). $\boldsymbol{r_q^a}=(x^a,y^a,z^a)$ are the actual positions of a quadrotor. $\boldsymbol{k}_p,\boldsymbol{k}_v,\boldsymbol{k}_i,\boldsymbol{k}_a$ are positive gains. The gain matrix is diagonal, and the corresponding diagonal element is denoted as $k_{(\bullet)i},\,i=x,y,z$. $t$ is the current time. $\delta t$ is a small positive value.

{Notably, the feedforward gains of accelerations are variable. They are used for reconciling the position-velocity tracking accuracy and agility.} From (\ref{eqn_ka}), the $k_{ai}$ will decrease if $|k_{pi}e_{pi}+k_{vi}e_{vi}|$ increases or $|\boldsymbol{\ddot r}_{qi}^r|$ decreases. The $\ddot r_{qi}^c$ tends to decrease the tracking errors and sacrifice agility. On the contrary, the $k_{ai}$ will increase. The $\ddot r_{qi}^c$ tends to increase agility and sacrifice the tracking {accuracy}. {The equation (\ref{eqn_ka}) also presents that if the current time is close to terminal time ($t \in (T-\delta t, T]$), the $\boldsymbol{k}_a$ will be $\boldsymbol{I}$. The other gains will be $0$.} The controller commands are taken entirely from  reference accelerations. This is to ensure the commanded attitude is consistent with the {surface's orientation.}

{The command accelerations are realized by the thrust and attitude of a quadrotor. They can be derived by the following equations:}
\begin{equation}
  \begin{split}
    &f=m(\boldsymbol{\ddot r_q^c}+g\boldsymbol{e}_3) \cdot \boldsymbol{Z}_q, \quad \psi=0\\
    &\phi=-\arcsin({a_y^c / \sqrt{{a_x^c}^2+{a_y^c}^2+({a_z^c+9.8})^2}})\\
    &\theta=\arctan{(a_x^c / ({a_z^c+9.8}))}.
  \end{split} \label{eqn_controlinput}
\end{equation}
{Where $f$ is thrust. $\phi,\theta and \psi$ is roll, pitch and yaw angle of a quadrotor respectively. The attitude loop controller is a cascade PID controller. It is provided by PX4-PixracerR15 autopilot.}

\subsection{Determination of Terminal States}
{The function \textit{GetTerminalStates($\boldsymbol{S}_p(T)$)} in Algorithm \ref{alg_timesearh} is used for determining the terminal states $\boldsymbol{S}_T$ of trajectories. In reality, $\boldsymbol{S}_T$ is determined by $\boldsymbol{S}_p(T)$, the conditions for successful perching and the control performance of the quadrotor. They are described by:}
\begin{gather}
    {\ddot y}_T=-9.8\sin{\phi_s}, \quad {\ddot z}_T=-9.8+9.8\cos{\phi_s} \label{eqn_deterterminalacc}\\
    {\dot y}_T={\dot y}_s(T)+\Delta V_{Y_s}\cos{\phi_s}-\Delta V_{Z_s}\sin{\phi_s} \label{eqn_deterterminalvely}\\
    {\dot z}_T={\dot z}_s(T)+\Delta V_{Y_s}\sin{\phi_s}+\Delta V_{Z_s}\cos{\phi_s} \label{eqn_deterterminalvelz}\\
    y_T=y_s(T)-l_{Z_s}\sin{\phi_s},\quad  z_T=z_s(T)+l_{Z_s}\cos{\phi_s}. \label{eqn_deterterminalpos}
\end{gather}

{Where $\Delta V_{Y_s}$ and $\Delta V_{Z_s}$ respectively are the desired tangential and normal relative velocities. They are represented in $Y_s-O_s-Z_s$ (see Fig.\ref{coordinatemodelforplancontrol}). $l_{Z_s}$ is the position offset along $Z_s$ axis.}

{From (\ref{eqn_deterterminalacc}), the terminal accelerations are determined by aligning the quadrotor $Z_q$ with surface $Z_s$. The thrust is set to the middle of its range.} From (\ref{eqn_deterterminalvely})-(\ref{eqn_deterterminalvelz}), the terminal velocities are determined by the surface velocities and the relative velocities $\Delta V_{Y_s},\Delta V_{Z_s} $. $\Delta V_{Y_s}$ and $\Delta V_{Z_s}$ are conditions for successful perching. {Equation} (\ref{eqn_deterterminalpos}) demonstrates that the terminal positions are determined by surface terminal positions and the normal offset distance $l_{Z_z}$. The distance is to tackle the lagging of attitude and position errors. {It is significantly determined by the lagging of the quadrotor's attitude\cite{thomas2016aggressive}.}

\section{Experiments, Results and Discussion}
In this section, {the tests of negative pressure and} two groups of experiments of perching on static and moving inclined surfaces are conducted. {A series of comparison experiments are also conducted to further show the mechanism of MD-Grippers.} {In subsection IV-A and B, the experiment settings are described.} In subsection \uppercase\expandafter{\romannumeral4}-C, the results of the experiments are demonstrated. Discussions are presented in subsection \uppercase\expandafter{\romannumeral4}-D.

\subsection{Tests of Negative Pressure with Inclined Surfaces}
\begin{figure}[!tb]
  \centering
  \includegraphics[width=3.0in]{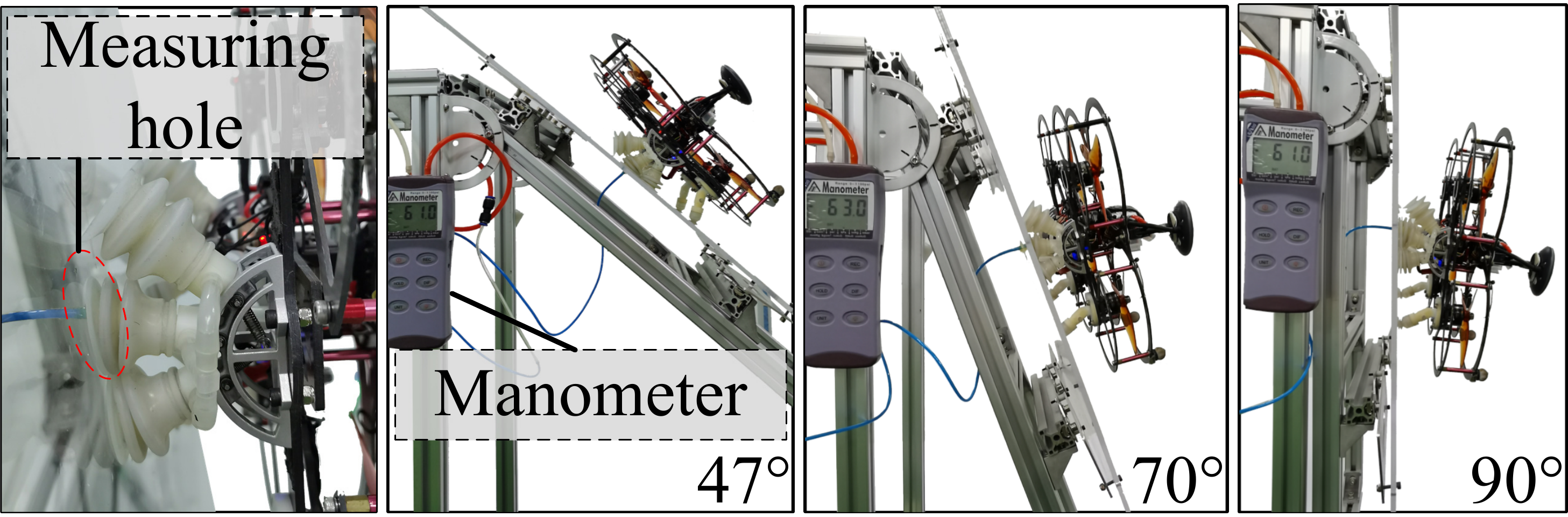}
  \caption{{The tests of negative pressure when the quadrotor perches on surfaces with different inclinations.}}
  \label{negativepressuretest}
\end{figure}
{The tests of negative pressure when the quadrotor perches on surfaces with different inclinations are shown in Fig.\ref{negativepressuretest}. We turn on the pneumatic system and put the aerial perching system on an inclined surface. Then, a slight pushing force acts on the quadrotor to ensure the self-sealing structure is opened and the perching is successful. The nominal maximum vacuum degree of the pump (SC3710PM) in the aerial perching system is $60kPa$. A measuring hole is drilled in the surface. A tube connects the hole and a manometer. In this way, we can measure the negative pressure if the hole is underneath the suction cup. We conduct five trials for surfaces with $47^\circ$, $70^\circ$ and $90^\circ$,  respectively. The results are plotted in Fig.\ref{negativepressureresult}.}

\subsection{Experiments of Perching on Static and Moving Inclined Surfaces}
\begin{figure}[!tb]
\centering
\includegraphics[width=3.0in]{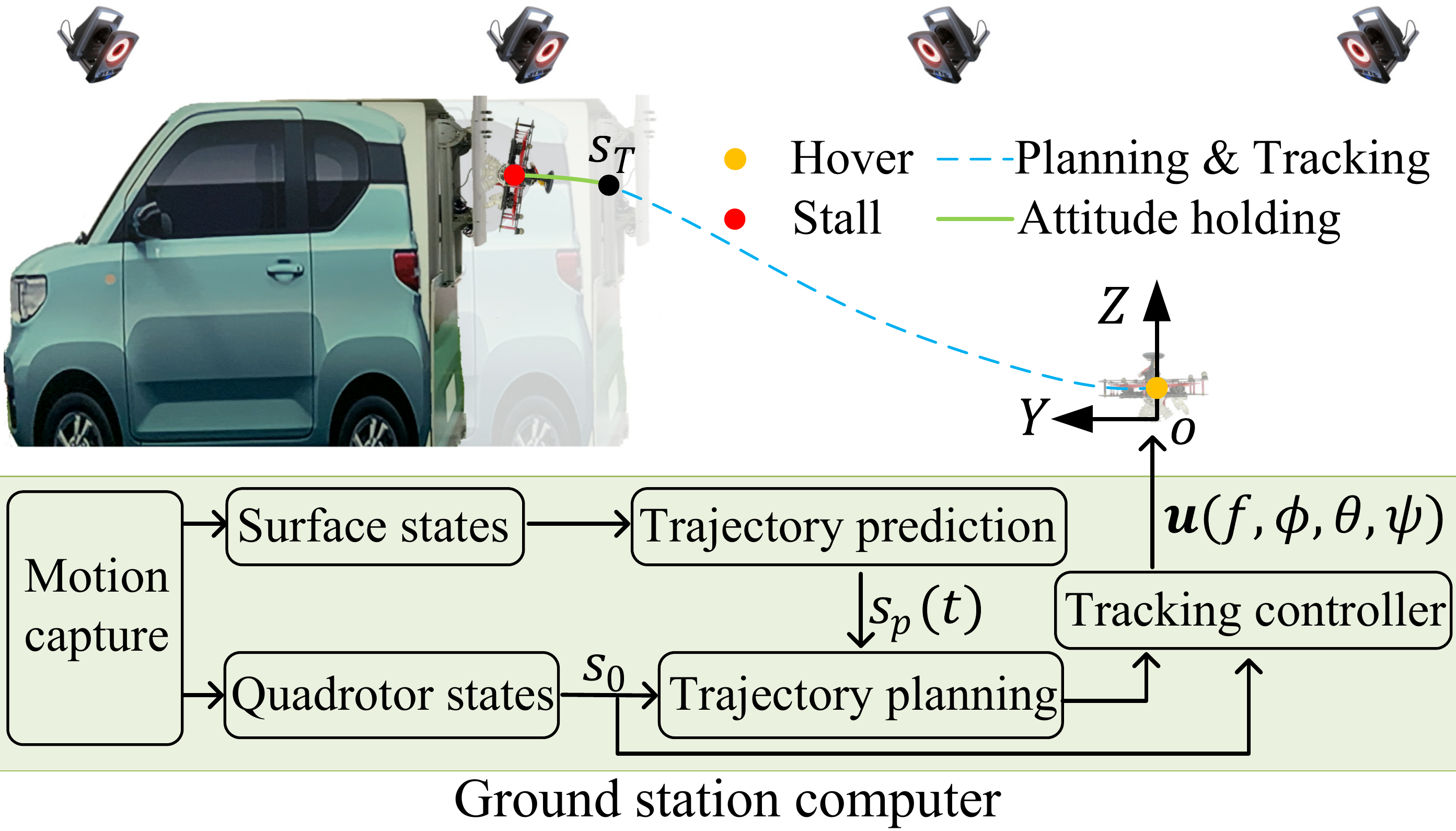}
\caption{The architecture of experiments for perching on a static or moving inclined surface.}
\label{experimentsetup}
\end{figure}
The architecture of perching on static inclined surfaces ({called static experiments}) is shown in Fig.\ref{experimentsetup}. An adjustable inclination surface is mounted on a car. The planning and tracking controlling algorithms are deployed on a ground station computer(Intel-i7-6800k CPU @ 3.4GHz, Ubuntu-18.04). {The pump runs all the time during experiments. Therefore, positive feedback can be formed when the impact force activates the suction cup.} {The control is run at 30$Hz$} and commands are sent to the quadrotor by a WIFI module. {A compass in a GPS module is also employed to obtain a more accurate yaw estimation.} The experiments are carried out under the Vicon motion capture system. Markers are stuck on the surface and quadrotor respectively to track their position and velocity (Fig.\ref{experimentsetup}). In the experiments, the car is stationary. {The quadrotor first hovers in the initial position. Then, the trajectory planning and tracking algorithms are executed.} At the end of the trajectories, the quadrotor is commanded to hold the attitude consistent with the surface. At last, the quadrotor is stalled if its height {doesn't change} for a period. {Three different inclination angles ($47^{\circ},70^{\circ}$ and $90^{\circ}$) of the surface are selected.} 20 trials are conducted for each case. {A group of comparison experiments for $70^{\circ}$ inclined surfaces  are also conducted by 20 trials. In these experiments, we replace MD-Grippers with two conventional suction cups (called Conv-Cups).} 

The architecture of perching on moving surfaces ({called moving experiments}) is similar to static experiments. In the experiments, the car is put in front of the quadrotor along the $Y$ axis. {Then, it is started and accelerated to its maximum speed by manual control.} The quadrotor {begins} the planning and tracking stage after the surface movement is detected. Perching is conducted while the car is accelerating. Then, the car is stopped after successful or failure perches are observed (We define the success as the quadrotor attaches on the surface over 2 seconds. {Other cases are regarded as failures}). The surface is set to $47^{\circ},70^{\circ}$ and $90^{\circ}$. {The car is driven forward (along $Y$ direction) and backward (opposite to $Y$ direction) respectively for every angle surface.} The comparison experiments using conventional cups for a $70^{\circ}$ surface are also conducted. Every configuration is tested by 10 trials. 

\subsection{Results}
\begin{figure}[!tb]
  \centering
  \includegraphics[width=3.0in]{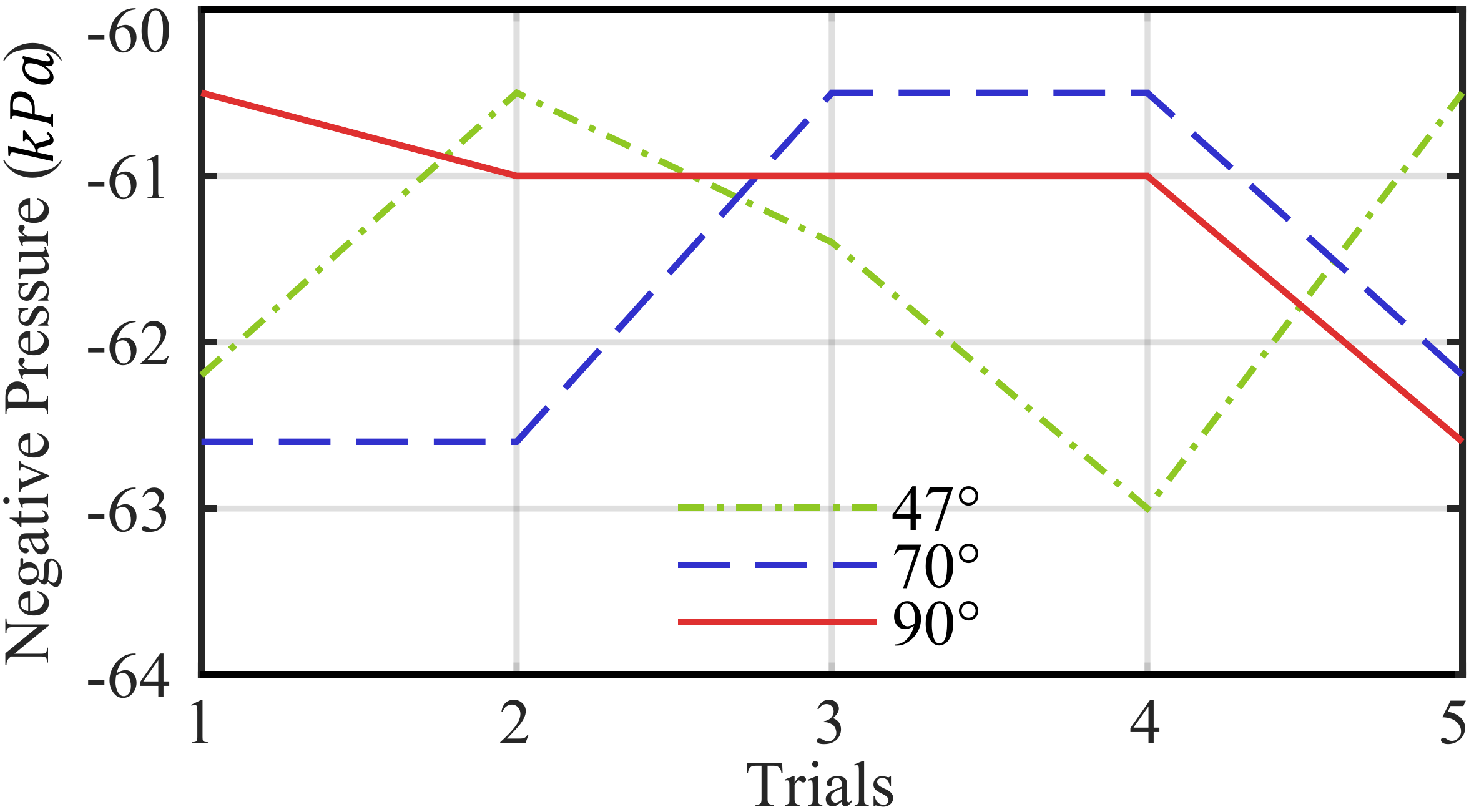}
  \caption{{The results of negative pressure tests.}}
  \label{negativepressureresult}
\end{figure}
{The results in Fig.\ref{negativepressureresult} demonstrate that the negative pressure does not change significantly for different inclinations. The average negative pressure is $-61.5kPa$.}

\begin{table}[!t]
  \centering
  \renewcommand{\arraystretch}{1.0}
  \caption{Results of Successful Perches on Static Surfaces}
  \label{table_staticsurfaces}
  \setlength{\tabcolsep}{1.0mm}{
  \begin{tabular}{@{}c|cccc@{}}
\toprule[1.2pt]
 \multirow{2}{*}{} &  \multicolumn{4}{c}{\makecell[c]{Inclination angle and Perching conditions \\($\Delta V_{Y_s}$m/s, \quad $\Delta V_{Z_s}$m/s, \quad $l_{Z_s}$m)}} \\ \cmidrule(l){2-5} 
  &  \makecell[c]{$47^{\circ}$ \\(0.3,-0.5,0.2)} & \makecell[c]{$70^{\circ}$ \\(0.3,-0.5,0.25) }  &  \makecell[c]{$90^{\circ}$ \\(0.3,-0.5,0.33) } & \makecell[c]{$70^{\circ}$ Conv-cups\\(0.3,-0.5,0.25)}   \\ \midrule[1pt]
 \makecell[c]{Success rate}  & 19/20     &  18/20  & 18/20 &  17/20            
 \\ \bottomrule[1.2pt]
\end{tabular}}
 \end{table}

 \begin{table}[!t]
  \centering
  \renewcommand{\arraystretch}{1.0}
  \caption{Results of Successful Perches on Surfaces Moving Forward}
  \label{table_positivemovingsurfaces}
  \setlength{\tabcolsep}{1.0mm}{
  \begin{tabular}{@{}c|cccc@{}}
\toprule[1.2pt]
 \multirow{2}{*}{} &  \multicolumn{4}{c}{\makecell[c]{Inclination angle and Perching conditions \\($\Delta V_{Y_s}$m/s, \quad $\Delta V_{Z_s}$m/s, \quad $l_{Z_s}$m)}} \\ \cmidrule(l){2-5} 
  &  \makecell[c]{$47^{\circ}$ \\(0.3,-0.2,0.07)} & \makecell[c]{$70^{\circ}$ \\(0.3,-0.2,0.23) }  &  \makecell[c]{$90^{\circ}$ \\(0.3,-0.1,0.15) } & \makecell[c]{$70^{\circ}$ Conv-cups\\(0.3,-0.2,0.23)}   \\ \midrule[1pt]
 \makecell[c]{Success rate}  & 9/10     &  7/10  & 7/10 &  2/10        \\\midrule
 \makecell[c]{Avg $\nu_{s}$ (m/s)}   & 0.96   & 0.93  & 1.07    & 1.18
 \\ \bottomrule[1.2pt]
\end{tabular}}
 \end{table}

 \begin{table}[!t]
  \centering
  \renewcommand{\arraystretch}{1.0}
\begin{threeparttable}[b] 
  \caption{Results of Successful Perches on Surfaces Moving Backward}
  \label{table_negativemovingsurfaces}
  \setlength{\tabcolsep}{1.0mm}{
  \begin{tabular}{@{}c|cccc@{}}
\toprule[1.2pt]
 \multirow{2}{*}{} &  \multicolumn{4}{c}{\makecell[c]{Inclination angle and Perching conditions \\($\Delta V_{Y_s}$m/s, \quad $\Delta V_{Z_s}$m/s, \quad $l_{Z_s}$m)}} \\ \cmidrule(l){2-5} 
  &  \makecell[c]{$47^{\circ}$ \\(0.3,-0.3,0.1)} & \makecell[c]{$70^{\circ}$ \\(0.3,-0.6,0.19) }  &  \makecell[c]{$90^{\circ}$ \\(0.3,-0.6,0.25) } & \makecell[c]{$70^{\circ}$ Conv-cups\\(0.3,-0.6,0.19)}   \\ \midrule[1pt]
 \makecell[c]{Success rate}  & 8/10     &  8/10  & 8/10 &  4/10        \\\midrule
 \makecell[c]{Avg $\nu_s$ (m/s)}   & -1.06   & -1.01  & -0.97    & -1.09
 \\ \bottomrule[1.2pt]
\end{tabular}}
\begin{tablenotes}
  \item[\quad  \quad  \quad \quad  1] Conv-cups means the experiment results using conventional cups 
\end{tablenotes}
\end{threeparttable}
 \end{table}
 {Table.\ref{table_staticsurfaces}-Table.\ref{table_negativemovingsurfaces} demonstrate the results of all successful perches. They are condition configurations, success rate and average surface velocity at impact(Avg $\nu_s$).} The results confirm that our system achieved reliable perching on static and moving inclined surfaces. {The success rate $\geq7/10$ and the maximum average $\nu_s$ reached $1.07$m/s.} The results also show the effectiveness of MD-Grippers and algorithm. As a contrast, the success rate of conventional cups decreases significantly ($2/10$ and $4/10$) in moving experiments.

\begin{figure}[!tb]
\centering
\includegraphics[width=3.5in]{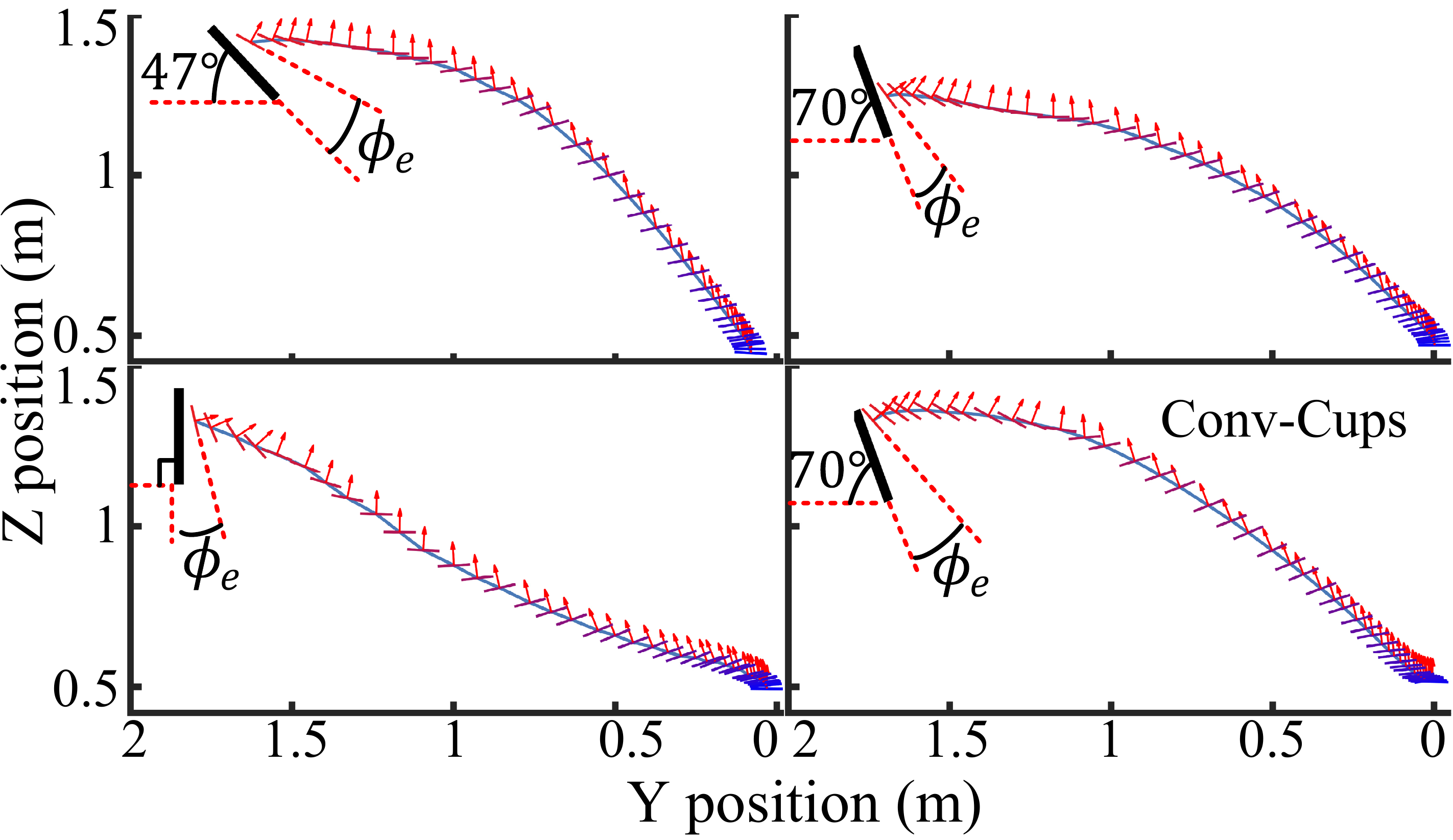}
\caption{The actual trajectories of the quadrotor perching on static inclined surfaces. The color from blue to red represents the passage of time. The short colored lines show the position and the orientation of the quadrotor. The arrow indicates the direction of the thrust. The short black line shows the position and the inclination of the surfaces. The impact angle errors are denoted as $\phi_e$.}
\label{statictrajectory}
\end{figure}
The trajectories of the quadrotor successfully perching on static inclined surfaces at $47^{\circ},70^{\circ}$ and $90^{\circ}$ are shown in Fig.\ref{statictrajectory}. These trajectories demonstrate the planner can generate aggressive trajectories to implement large attitude maneuvers. {Further, the angle errors when the grippers impact the surfaces are also presented (denoted as $\phi_e$).} They are also the angle errors needed to be accommodated by grippers. 

\begin{figure}[!tb]
  \centering
  \includegraphics[width=3.5in]{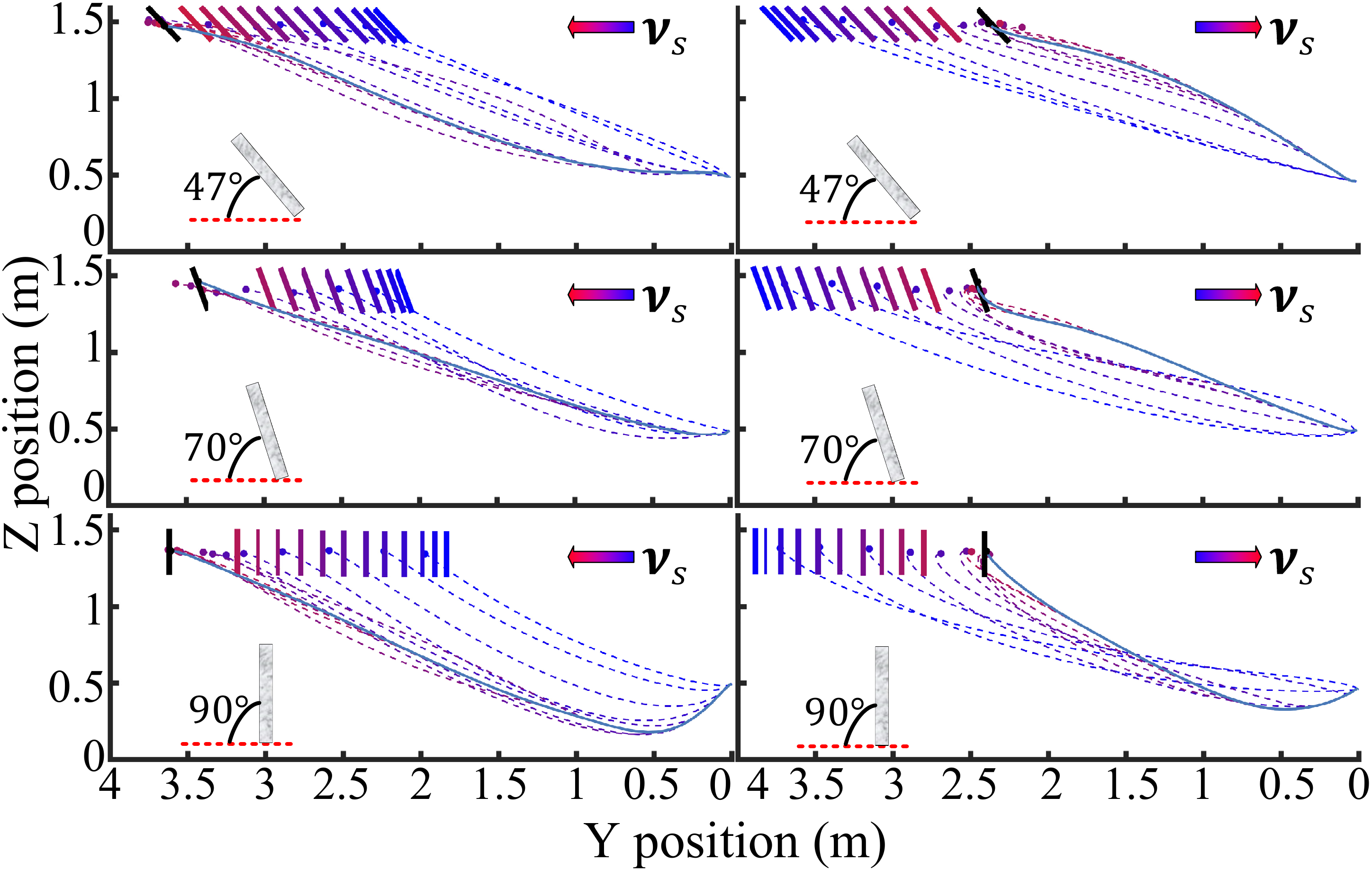}
  \caption{The planning and actual trajectories of the quadrotor perching on moving inclined surfaces. The solid colored lines show the position and the inclination of the surfaces at different times. The dashed colored lines show the planned trajectory based on the predicted rendezvous (denoted as the colored dots). The blue solid curves are the actual trajectories.}
  \label{movingtrajectory}
  \end{figure}
  \begin{figure}[!tb]
  \centering
  \includegraphics[width=3.0in]{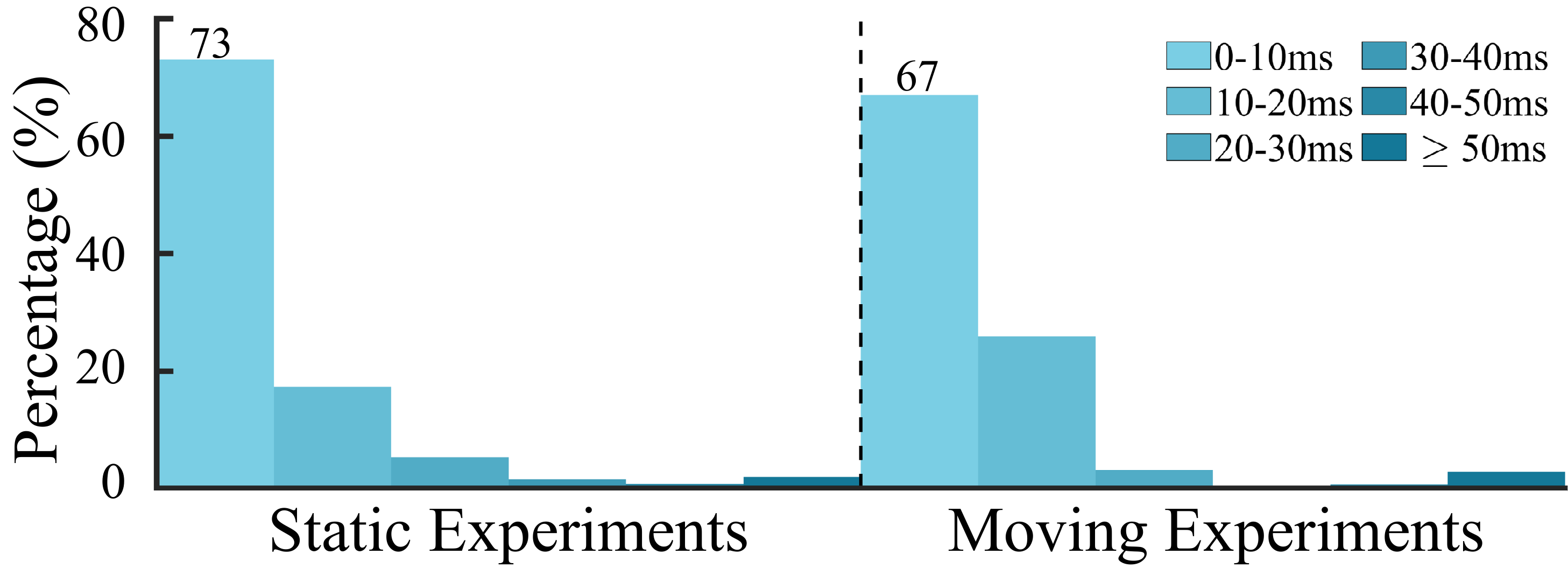}
  \caption{The distribution of the planning algorithm solution time.}
  \label{solutiontimecost}
  \end{figure}
  The real-time planned and actual trajectories of the quadrotor perching on moving surfaces are shown in Fig.\ref{movingtrajectory}. We can know the predicted rendezvous (colored dots) gradually converges when the surface's velocity $\nu_s$ tends to be constant. {The trajectories can also be updated as surfaces move.} The solution time of the planning algorithm is  distributed as Fig.\ref{solutiontimecost}. It suggests $73\%$ and $67\%$ of the solution time are less than $10ms$ {for static and moving perching respectively.} The computational time is smaller than that in \cite{hu2019time}. Consequently, trajectories can be planned in real-time to adapt to the movement of surfaces. In the cases with long process time, the last generated trajectories will be adopted such that the perching will not stop. An image sequence of perching on a surface moving forward is shown in Fig.\ref{compareandfailurepics} (d). More details are shown in the submitted video. 

\begin{figure}[!tb]
\centering
\includegraphics[width=3.0in]{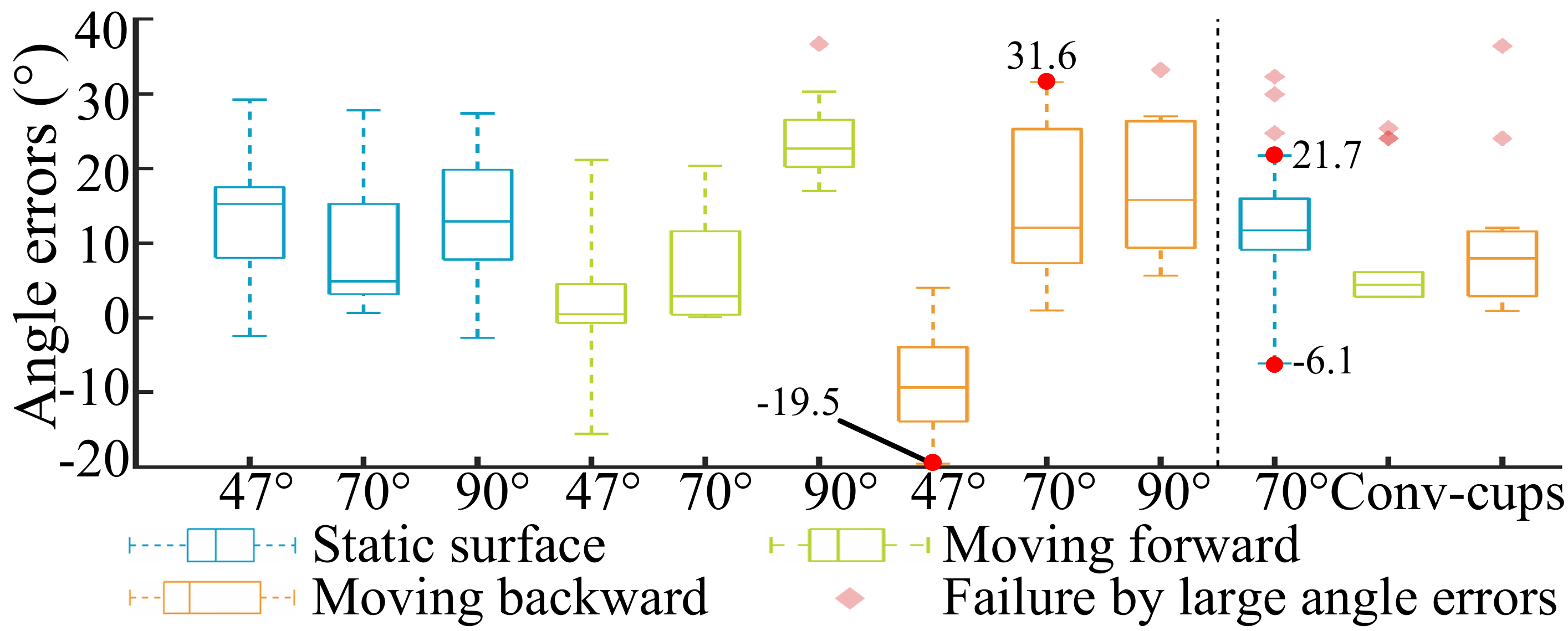}
\caption{The distribution of angle errors $\phi_e$ for all successful perches. Some failures due to angle errors exceeding upper limits. The blue, green and yellow boxes indicate successful perches on static surfaces, surfaces moving forward and surfaces moving backward respectively. The right part is the results of comparison experiments for $70^\circ$ surfaces. The red diamonds represent the failure caused by large angle errors.}
\label{angleerrorsdistri}
\end{figure}
\begin{figure}[!tb]
\centering
\includegraphics[width=3.2in]{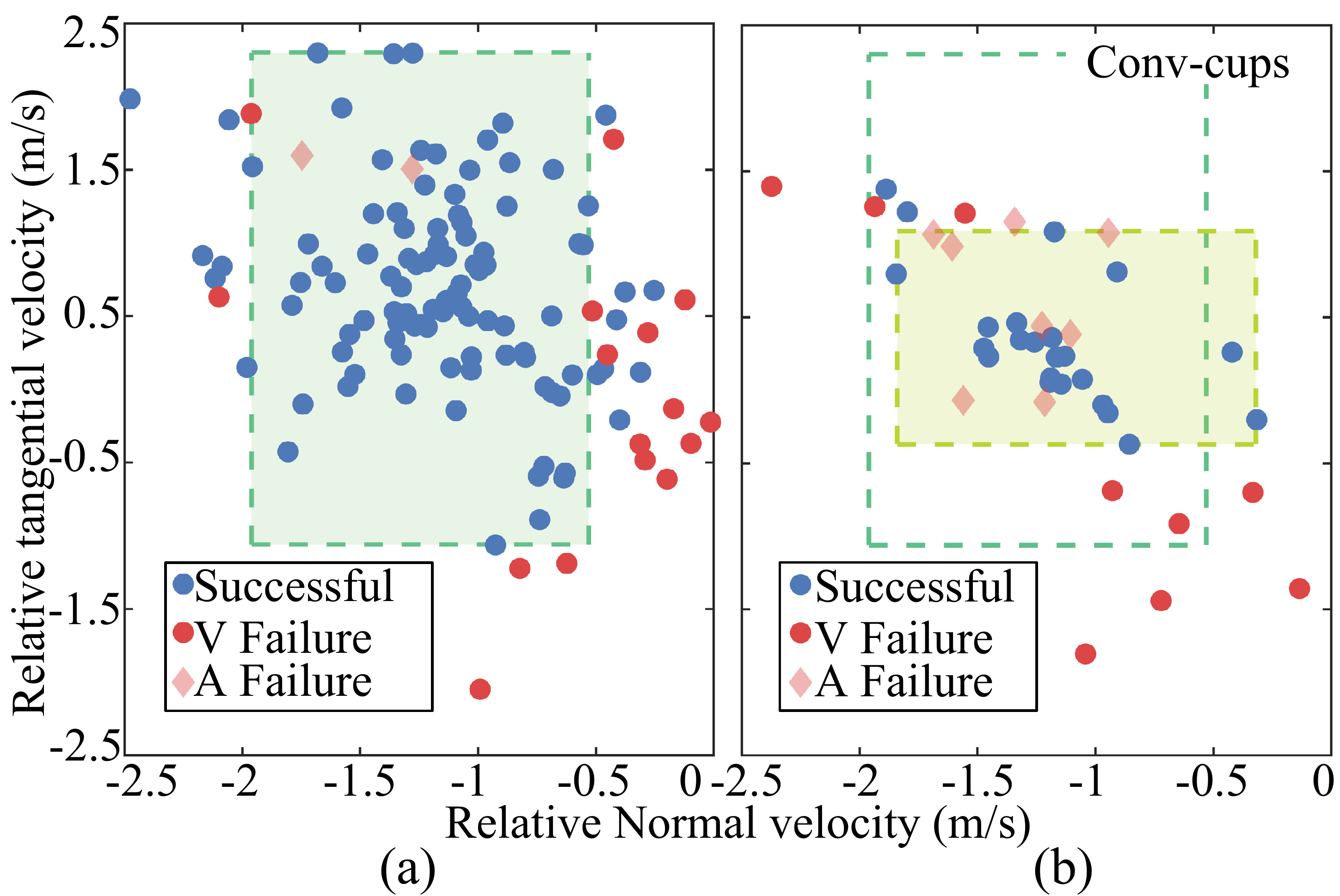}
\caption{The distribution of actual impact velocities relative to a surface in tangential and normal directions. (a) The results of the experiments using MD-Grippers. (b) The results of the experiments using the Conv-Cups. The blue dots indicate the successful perches. The red dots indicate the failures caused by improper relative velocities ({called} V Failure). The light red diamonds indicate the failures caused by large angle errors ({called} A Failure). The green area is the feasible region of impact velocities for MD-Grippers and the yellow area is the feasible region of Conv-Cups.}
\label{collisionveldistribution_horizon}
\end{figure}
The distribution of impact angle errors $\phi_e$ of all successful perches are plotted in Fig.\ref{angleerrorsdistri}. The upper and lower limits of $\phi_e$ of MD-Gripper are ($-19.5^\circ$, $31.6^\circ$). {The range for Conv-Cups are ($-6.1^\circ$, $21.7^\circ$).} The perches failed once $\phi_e$ exceeds the upper limits (plotted as the light red diamonds). {Therefore, the upper and lower limits determine a range. It is considered the angle error range that the gripper can accommodate.} Obviously, the MD-Gripper has larger adaptability to angle errors than Conv-Cup. 

The impact velocities of the quadrotor relative to surfaces are plotted in Fig.\ref{collisionveldistribution_horizon}. The results of the experiments using MD-Grippers and the Conv-Cups are plotted in Fig.\ref{collisionveldistribution_horizon} (a) and (b) respectively. The horizontal and vertical axes are the actual $\Delta V_{Z_s}$ and $\Delta V_{Y_s}$. The successful perches are plotted as blue dots and the failures are plotted as red dots or light red diamonds. {Herein, if the angle errors of one failure are out of the tolerant range, we consider large angle errors cause the failure. It is plotted as a light red diamond (also shown in Fig.\ref{angleerrorsdistri}).} Other failures are considered to be caused by improper velocities and plotted as red dots. 

The green or yellow area is the largest rectangle that contains no red dots and has at least one blue dot on each side. The {rectangle's boundaries} determine the tolerant range to the impact tangential and normal velocities. The velocities located in the rectangle are considered to meet the perching conditions. Apparently, the green rectangle is significantly larger than the yellow one. That means the MD-Gripper has larger adaptability to velocities than Conv-Cup.
  
\subsection{Discussion}
\begin{figure}[!tb]
\centering
\includegraphics[width=3.0in]{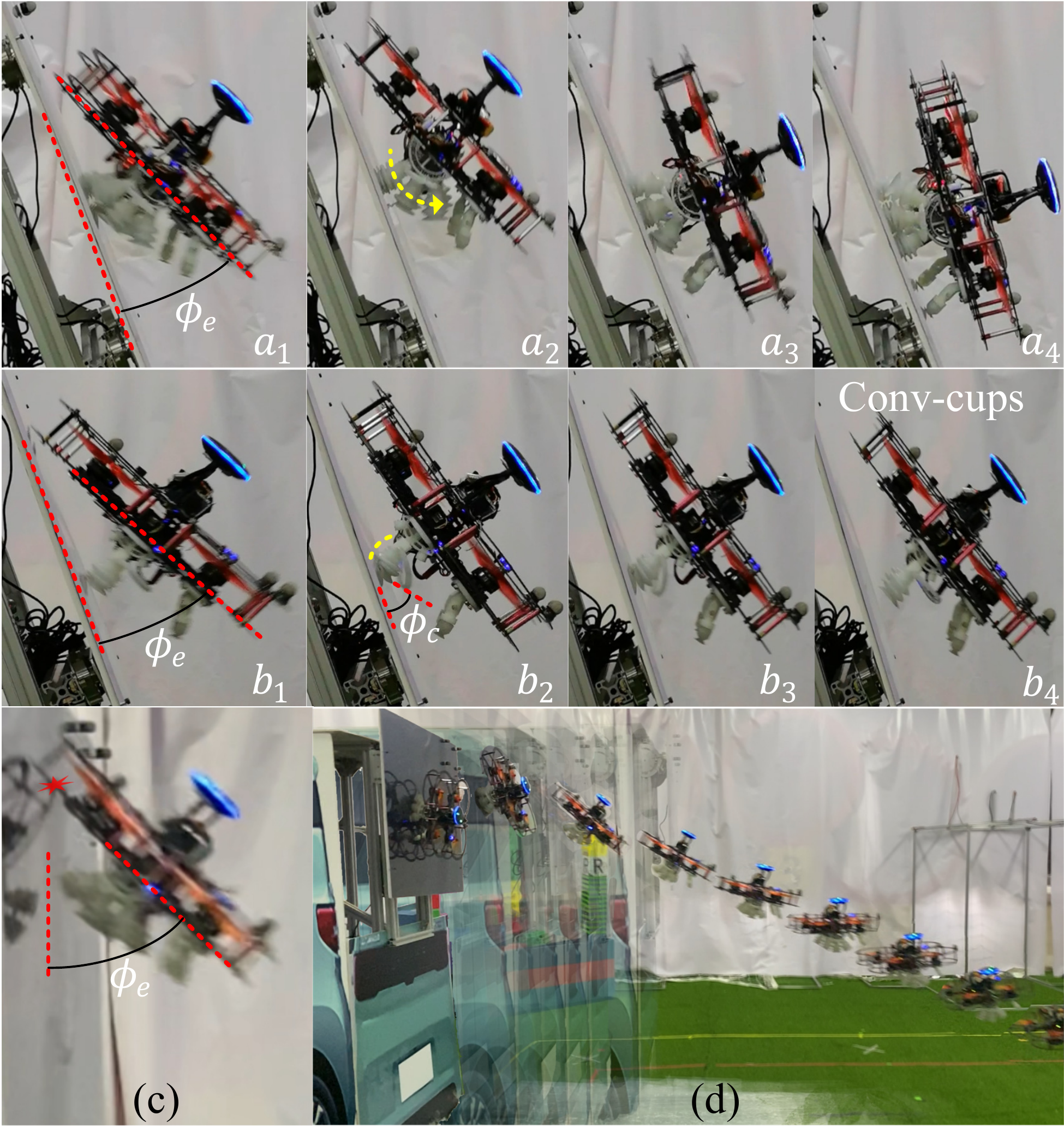}
\caption{The snapshots of some experiments. $a_1$-$a_4$ shows the adaptability of MD-gripper to angle error $\phi_e$. $b_1$-$b_4$ shows the failure process of one comparison experiment using conventional cups caused by angle error $\phi_e$. (c) A large angle error causes the collision between the airframe and the surface. (d) The strobe images of a successful perch on a vertical surface moving forward.}
\label{compareandfailurepics}
\end{figure}
{The negative pressure in the suction cups does not change significantly for inclined surfaces. This is because of the positive feedback mechanism in the self-sealing suction cup. The negative pressure will approximately equal the maximum vacuum degree of the pump. In our experiments, the average negative pressure $P_G$ is $-61.5kPa$. The radius of the suction cup is $R=0.015m$. The adhesion force for two grippers is $F_G=-2\pi R^2P_G \approx 86.9N$. The friction coefficient $\mu$ between the suction cup and glass is approximately $0.3$\cite{liu2006analytical}. The maximum friction force $F_f=\mu F_G\approx 26.1N$. The weight of our perching system is $9.26N$. Therefore, the load capacity provided by the pneumatic system is sufficient for the perching system.}

The larger angle errors adaptability of the MD-Gripper benefits from the employment of the SS-Cup in multi-directions. First, the angle $\alpha_w$ between two adjacent SS-Cups increases the adaptability to $\phi_e$. Second, the wheel mechanism is adaptive to the tangential velocity. The process is shown in Fig.\ref{compareandfailurepics} ($a_1$-$a_4$). There is an angle error $\phi_e$ when the grippers impact the surface (see $a_1$) and  the tangential velocity $\Delta V_{Y_s} >0$. Therefore, the upper SS-Cup in MD-Gripper is {employed as the gripper rotates. ${\boldsymbol{n}_c \cdot \boldsymbol{\nu} \geq 0}$ in (\ref{eqn_torquevelocity}) is also satisfied for the upper cup. It facilitates the sealing of the cup (see $a_2$). However, for Conv-Cups, the tangential velocity bends the cup due to ${\boldsymbol{n}_c \cdot \boldsymbol{\nu} < 0}$. That increases the angle error $\phi_c$ (see Fig.\ref{cupmodule}) and destroys the sealing of the Conv-Cups(Fig.\ref{compareandfailurepics} ($b_2$)).} {In reality, MD-Gripper can accommodate the tangential velocity in different directions whether $\phi_e\geq 0$ or $\phi_e <0$. That enlarged the adaptability to the tangential velocity (Detailed in case1-2 of the submitted video). Therefore, the green rectangle in Fig.\ref{collisionveldistribution_horizon} is significantly larger than the yellow one in the tangential direction.}

In moving experiments, the angle and velocity errors will be more significant. They are likely to exceed the adaptability of Conv-Cups. {That reduces the success rate significantly.} Comparatively, MD-Gripper has larger adaptability to angle errors and velocities. {Furthermore, the larger adaptability to velocities makes a velocity range feasible. The range allows the quadrotor to translate a distance while keeping its attitude aligned with the surface. This enables the quadrotor to adapt to certain position errors.} Therefore, the success rate keeps higher. Reliable perching on moving surfaces is achieved.

It should be pointed out that the adaptability of MD-Gripper is also limited. In reality, if $\phi_e$ is too large, the collision between the airframe and the surface will occur. This usually leads to failure (Fig.\ref{compareandfailurepics} (c)). Besides, the failures will happen if $\Delta V_{Z_s}$ exceeds the right boundary or the left boundary in Fig.\ref{collisionveldistribution_horizon} (a). {This is because the small impact velocities cannot provide enough contact force to seal the SS-cup and open the trigger cup. Excessive impact velocities will cause the quadrotor to bounce off the surface.} Moreover, if $\Delta V_{Y_s}$ is less than the lower boundary, the suction cup would be pealed off by downward momentum. All these failures are plotted as red dots in Fig.\ref{collisionveldistribution_horizon} (a).

{Additionally, the perching conditions $\Delta V_{Y_s},\Delta V_{Z_s}$ and $l_{Y_s}$ in our experiments are set by pre-experiments.} From Fig.\ref{collisionveldistribution_horizon}, the actual $\Delta V_{Y_s},\Delta V_{Z_s}$ are dispersed in a large range and there are significant terminal errors in reality. In our system, the multiple suction cup design would disturb the flow from rotors. This is also detrimental to the control accuracy. Fortunately, the large adaptability to angle errors and velocities facilitates the selection of these parameters. The obtained perching condition (rectangle in Fig.\ref{collisionveldistribution_horizon}(a)) could also be a reference in the future. A controller considering disturbance rejection and asynchronization  between thrust and attitude \cite{8823973} might be necessary to reduce the attitude and velocity errors.

{Finally, we have not considered strategies to make the quadrotor take off from inclined surfaces. To achieve this ability, the pneumatic system should be improved to make the air enter suction cups by using valves. Furthermore, a controller that can resist and adapt to the disturbances from the surface is required during take-off. This will be another research topic in further.}

\section{Conclusion}
In this article, we designed a gripper and a real-time trajectory planner to enable a quadrotor to perch on moving inclined surfaces. {The gripper comprises  multiple self-sealing suction cups in different directions. It can adapt to attitude errors and tangential velocities in different directions during perching.} The multimodal dynamic time-domain search method is developed to obtain a time-optimal trajectory. The trajectories respecting motor lift constraints can be generated in real-time to adapt to the surface's movement. {The extensive experiments of perch on static and moving inclined (up to $90^\circ$) surfaces verify the effectiveness of the proposed system.} The comparative test results demonstrate our proposed gripper has larger adaptability to angle errors and tangential velocities. The planning algorithm has high real-time performance. {The quadrotor can perch on moving inclined surfaces (maximum average velocity $\geq 1.07$m/s) with a high success rate ($\geq 15/20$). It outperforms the conventional suction cup significantly ($15/20 > 6/20$).} The results suggest that the designed gripper and algorithm can be employed to perch on moving objects with smooth surfaces. {The technique can be exploited in air-ground cooperative tasks to extend the operation time and range of quadrotors.} Another promising extension to our system is to grasp moving objects with online trajectory planning.

As a follow-up, further work can be focused on leveraging the onboard sensors to enable perching on a vehicle {outdoors.} {An observation-execution collaborative framework will be studied. The system's adaptability to surface position errors will also be analyzed and exploited for outdoor applications.} {Other configurations of the devised self-sealing suction cup could also be explored to reduce disturbance to the system.} The strategy to {perch} on a vehicle changing its heading would also be researched. {A pneumatic system and disturbances rejection controller shall also be developed to enable quadrotors to take off from inclined surfaces.}
\ifCLASSOPTIONcaptionsoff
  \newpage
\fi
\bibliographystyle{IEEEtran}
\bibliography{IEEEabrv,aggcupreference}

%

\vspace{-50 pt} 
\begin{IEEEbiography}[{\includegraphics[width=1in,height=1.2in,clip]{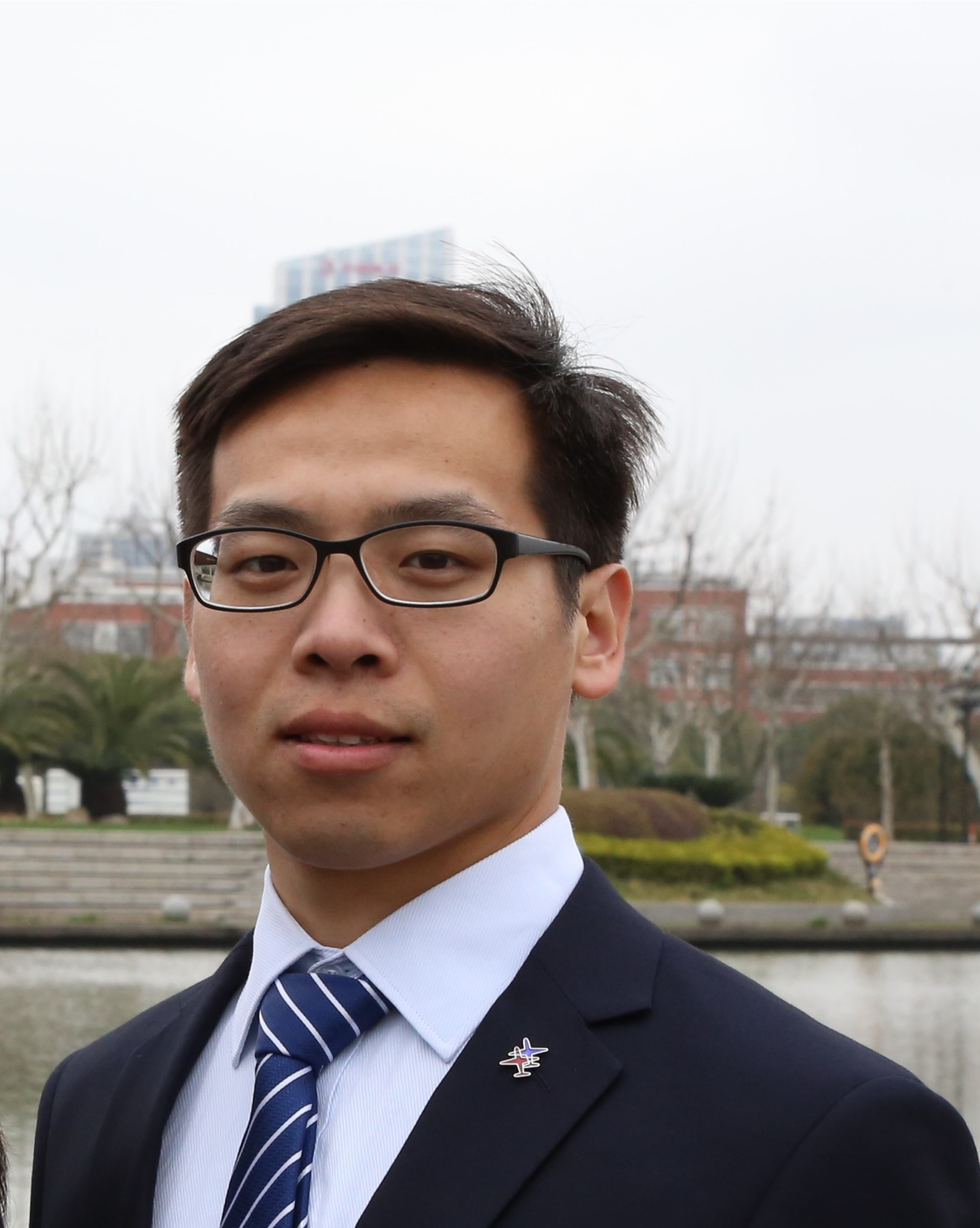}}]{Sensen Liu} received the B.S. degree in mechanical engineering from Tongji University, Shanghai, China, in 2016. He is currently a Ph.D. candidate in the School of mechanical and engineering, Shanghai Jiao Tong University, China. His research focuses on aerial manipulation, gripper design, planning and control of unmanned aerial vehicles manipulation system. 
\end{IEEEbiography}
\vspace{-50 pt} 
\begin{IEEEbiography}[{\includegraphics[width=1in,height=1.2in,clip]{Fig-bio-zywang-eps-converted-to}}]{Zhaoying Wang} received the B.E. degree in mechanical engineering from Wuhan University, Wuhan, China, in 2019. He is now a Ph.D. candidate at the State Key Laboratory of Mechanical System and Vibration, Shanghai Jiao Tong University. His research interest is multi-agent cooperative system and visual positioning of micro aerial vehicles.
\end{IEEEbiography}
\vspace{-50 pt}
\begin{IEEEbiography}[{\includegraphics[width=1in,height=1.2in,clip]{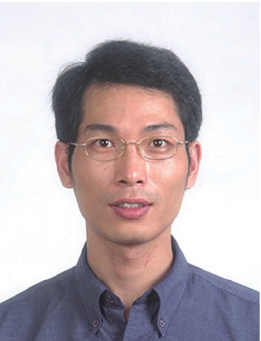}}]{Xinjun Sheng} received the B.Sc., M.Sc.,and Ph.D. degrees in mechanical engineering from Shanghai Jiao Tong University, Shanghai, China, in 2000, 2003, and 2014, respectively. He is currently a Professor in the School of Mechanical Engineering, Shanghai Jiao Tong University. His current research interests include robotics, and bio-mechatronics. Dr. Sheng is a Member of the IEEERAS, the IEEEEMBS,and the IEEEIES.
\end{IEEEbiography} 
\vspace{-500 pt} 
\begin{IEEEbiography}[{\includegraphics[width=1in,height=1.2in,clip]{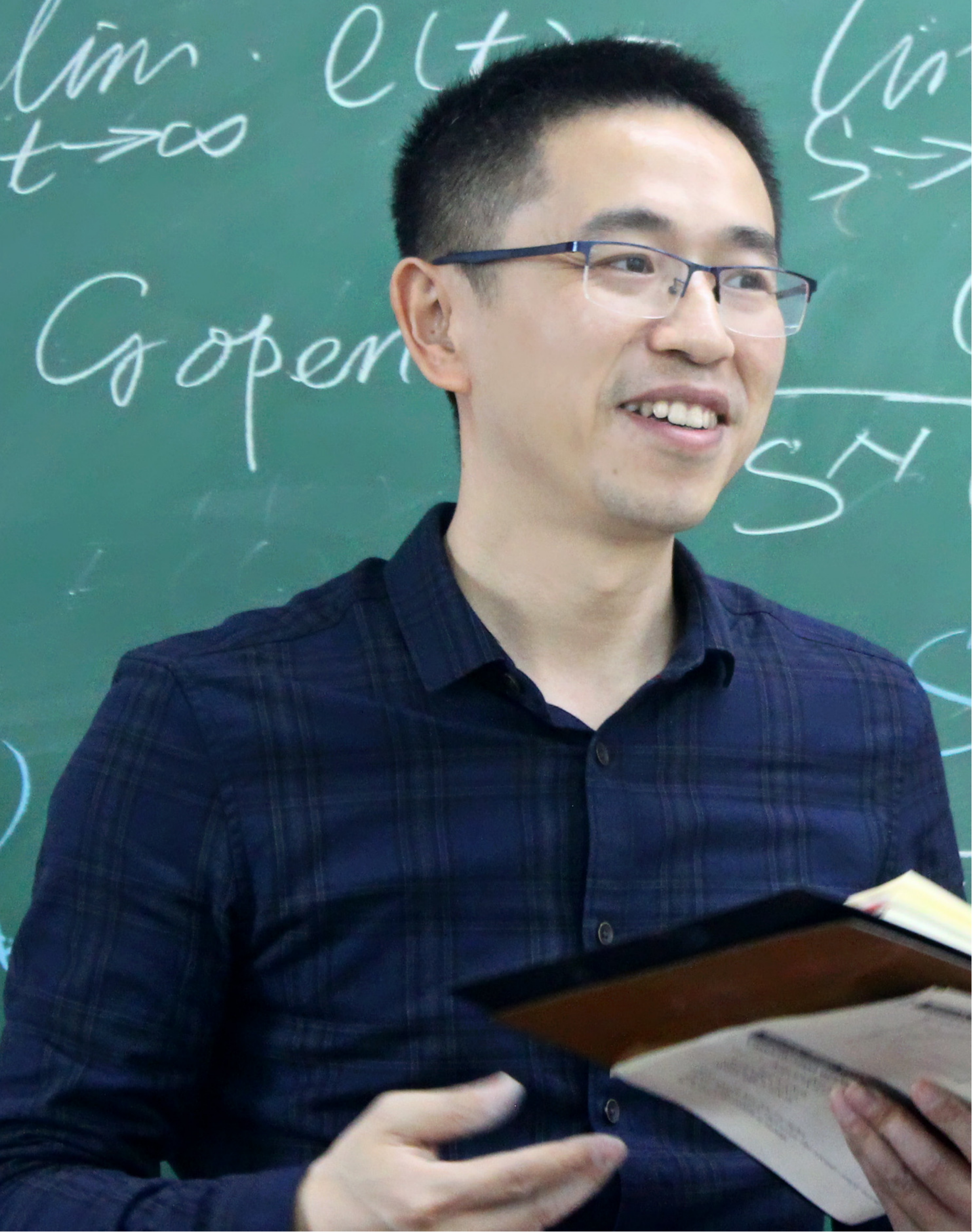}}]{Wei Dong} received the B.S. degree and Ph.D. degree in mechanical engineering from Shanghai Jiao Tong University, Shanghai, China, in 2009 and 2015, respectively. He is currently an associate professor in the School of Mechanical Engineering, Shanghai Jiao Tong University. His research interests include cooperation, perception and agile control of unmanned system.
\end{IEEEbiography}




\end{document}